# Generating Realistic Safety-Critical Scenarios for Vehicle-Pedestrian Interactions

Qingwen Pu[a], Kun Xie[a,*], Yuan Zhu[b], and Guocong Zhai[c]

[a] *Transportation Informatics Lab, Department of Civil and Environmental Engineering, Old Dominion University, Norfolk, VA 23529, United States*
[b] *Inner Mongolia Center for Transportation Research, Inner Mongolia University, Rm A357A, Transportation Building, Inner Mongolia University South Campus, 49 S Xilin Rd, Hohhot, Inner Mongolia, 010020, China*
[c] *School of Transportation and Logistics, National Engineering Laboratory of Integrated Transportation Big Data Application Technology, National and Local Joint Engineering Research Center of Integrated Transportation Intelligence, Southwest Jiaotong University, Chengdu 611756, China*
[*] *Corresponding Author; Email: kxie@odu.edu*

## Abstract

Automated driving system (ADS) deployment requires rigorous validation across safety-critical vehicle-pedestrian interactions, yet real-world datasets rarely capture high-risk scenarios while simulation platforms lack realistic behavior. In response, this study proposes a three-stage framework that combines real-world grounding with adaptive simulation to generate behaviorally realistic safssety-critical scenarios at scale. Stage 1 pre-trains multi-agent state-space Transformer-enhanced DDPG (MA-SST-DDPG) agents on real-world safety-critical data to learn human-like interactive evasive behaviors through data-driven learning. Stage 2 deploys pre-trained multi-agents in CARLA for online reinforcement learning to generalize across diverse scenarios, integrating real-world knowledge with simulation experience to produce a refined MA-SST-DDPG model. Stage 3 uses CARLA with the refined model to generate over 198,000 high-resolution interaction episodes from eight intersection scenarios, culminating in the Vehicle–Pedestrian Safety-Critical Interaction (VPSCI) dataset. The Refined MA-SST-DDPG model outperformed baseline methods in reproducing realistic evasive behaviors, achieving the lowest trajectory errors (ADE = 0.072 m, FDE = 0.142 m). Statistical comparison confirmed distributional equivalence between the generated and real-world data in both conflict severity and behavioral response. A Turing test confirmed that the three-stage framework generated evasive behaviors were indistinguishable from real-world interactions. The generated data revealed realistic behavioral patterns: conflict rates rose with speed, and pedestrian yielding increased with vehicle proximity and speed—trends matching real-world observations. These results demonstrate the framework's effectiveness in producing high-fidelity safety-critical data, offering valuable sources for the development of ADS and simulation-based safety evaluations.

*Keywords*: Safety-critical scenario synthesis, Online reinforcement learning, Human-like evasive behavior, CARLA Simulation, Interactive decision-making

## 1. Introduction

The development of automated driving systems (ADS) hinges critically on their ability to safely navigate complex and dynamic urban environments. Among the most challenging and consequential situations are interactions between vehicles and pedestrians at urban intersections (Ni *et al.* 2016), where the margin for error is minimal and the stakes are often life-threatening. Safety-critical scenarios between vehicles and pedestrians provide important safety implications, but they are rare and hard to capture with traditional methods (Zhou *et al.* 2025). The lack of detailed safety-critical data limits the development and evaluation of ADS, which require large-scale, diverse, and realistic interaction scenarios to ensure performance (Imaseki et al. 2023, Yang et al. 2024). Therefore, generating datasets that capture safety-critical scenarios involving vehicle-pedestrian interactions is crucial for enabling ADS to learn complex intersection dynamics, develop effective crash avoidance strategies, and ultimately enhance pedestrian safety (Zhang *et al.* 2018, Xu *et al.* 2024).

Existing public datasets provide important resources but remain insufficient for comprehensive ADS safety validation. Real-world datasets such as NGSIM (Alexiadis et al. 2004), INTERACTION (Zhan et al. 2019), and Waymo Open Motion (Sun et al. 2020) capture authentic human behaviors with precise trajectory data. However, they lack sufficient safety-critical near-miss scenarios (Rempe *et al.* 2022). Even specialized datasets like inD (Bock et al. 2020) and HDI (Pu *et al.* 2025b) that focus on urban intersections rarely capture the high-risk collision-avoidance behaviors essential for training ADS. To overcome this data scarcity, researchers have turned to simulation-based datasets using platforms like CARLA (Dosovitskiy et al. 2017), Virtual KITTI (Gaidon *et al.* 2016), and SYNTHIA (Ros *et al.* 2016), which enable large-scale generation of safety-critical scenarios. However, the driving and pedestrian behaviors in these simulators are primarily governed by rule-based controllers and simplified kinematic models, including predefined car-following and lane-changing heuristics implemented in CARLA's Traffic Manager (Shi *et al.* 2023). These behaviors follow fixed mathematical rules rather than adaptive decision-making, limiting their ability to reproduce realistic, interactive, and evasive responses in safety-critical vehicle-pedestrian encounters (Suo *et al.* 2021). This creates a challenge: real-world methods lack sufficient safety-critical data while simulation methods lack behavioral realism. As a result, hybrid approaches that effectively integrate both domains are needed (Suzuki *et al.* 2017, Pradana *et al.* 2022).

To address the scarcity of safety-critical data, recent studies have proposed hybrid frameworks that first learn human-like behaviors from real-world data and then generate diverse interaction scenarios in simulation. For instance, BITS (Xu *et al.* 2023), AADS (Li *et al.* 2019), and RoboTron-Sim (Xiao *et al.* 2025) use imitation learning or behavior cloning to initialize realistic agent policies before large-scale scenario synthesis. Others have incorporated digital twin systems (Alnowaiser and Ahmed 2024), or critical case generation (Akhauri *et al.* 2020, Song *et al.* 2025) to reduce the simulation–reality gap. However, these hybrid approaches share a fundamental limitation: they treat simulation as a deployment environment rather than a learning environment. Once pre-trained on real-world data, agent policies remain frozen during simulation-based generation, unable to adapt to new scenarios or interaction patterns (Suo *et al.* 2021). Meanwhile, this "learning-then-generating" paradigm relies exclusively on real-world data to capture safety-critical behaviors, but real-world datasets are too scarce to learn diverse safety-critical interactions (Ding *et al.* 2023). Consequently, generated scenarios exhibit limited diversity beyond the pre-training data, and agents demonstrate brittle decision-making under shifted simulation conditions (Zhao *et al.* 2022). Without online learning to refine policies through simulation interaction, these frameworks cannot exploit simulation's scalability to alleviate limitations caused by scarce real-world interaction data.

This study addresses these challenges through a three-stage framework that integrates real-world behavioral grounding with large-scale, simulation-driven scenario generation. Stage 1 pre-trains multi-agent state-space Transformer-enhanced deep deterministic policy gradient (MA-SST-DDPG) agents on 336 real-world safety-critical interactions from the HDI dataset (9,443 time steps, CurvTTC < 5 s) to provide an initial policy representation for collision avoidance. Stage 2 deploys these agents in CARLA's Town10 for online reinforcement learning across diverse intersection layouts. When CurvTTC falls below 5 s, MA-SST-DDPG overrides CARLA's autopilot to execute evasive maneuvers. Through iterative simulation interaction, agents integrate real-world knowledge with simulation experience, yielding a refined model that generalizes beyond the initial training distribution. Stage 3 applies this refined model to generate 198,157 interaction episodes across eight scenarios (two locations, four crossing directions each), capturing 3,603 km of vehicle and 2,625 km of pedestrian trajectories. The Vehicle–Pedestrian Safety-Critical Interaction (VPSCI) Dataset is publicly available at https://github.com/Qpu523/VPSCI-Dataset. The main contributions of this study include:

• Propose a three-stage data generation framework featuring a hybrid reinforcement learning approach that integrates data-driven pre-training with simulation-driven online learning, addressing the scarcity of real-world data and the lack of behavioral realism in simulation-based methods.

• Construct the VPSCI dataset—a large-scale, high-resolution collection of over 198,000 safety-critical vehicle–pedestrian interaction episodes across diverse intersection scenarios, validated through a Turing test to ensure behavioral realism.

• Demonstrate the use of three-stage framework generated data for performing safety analysis, enabling quantitative evaluation of conflict rates and behavioral risk patterns across diverse interaction scenarios.

## 2. Literature Review

### 2.1. Real-World and Simulation-Generated Safety-Critical Data

The safety of ADS has become a major research focus worldwide (Quispe *et al.* 2022). Early large-scale tests investigated vehicle–vehicle scenarios such as merging, exiting, car-following, and lane-changing, applying continuous acceleration profiles to assess the safety limits of autonomous vehicles (Ali 2025). These tests set benchmarks for longitudinal control and collision avoidance in structured traffic (Li *et al.* 2021). As AVs are introduced into urban environments, safety evaluation must also cover mixed traffic with vulnerable road users, especially pedestrians (Alozi and Hussein 2022, Harkin *et al.* 2024).

For vehicle–pedestrian interactions, comprehensive datasets require diverse intersection layouts, varied traffic densities, and realistic evasive actions (Pu *et al.* 2026d). Right-turn scenarios merit particular attention as they lack signal restrictions, yielding higher vehicle-pedestrian conflict frequencies than signal-controlled left turns (Feng *et al.* 2025). Such data are important for behavior modeling and safety evaluation (Feng *et al.* 2025). However, existing real-world and simulation-based datasets remain insufficient, exhibiting distinct limitations detailed in Table 1.

Real-world datasets record accurate movement data but rarely capture enough near-miss events for reliable crash-avoidance studies (Dulac-Arnold *et al.* 2019). Highway datasets like NGSIM (Alexiadis et al. 2004) and HighD (Krajewski et al. 2018) provide detailed vehicle trajectories on highways and arterials, supporting research in traffic flow and car-following. However, they exclude pedestrian movements and focus mainly on longitudinal interactions under normal conditions (He 2017, Zhan *et al.* 2019). More urban-focused datasets, such as INTERACTION (Zhan et al. 2019) and inD (Bock et al. 2020), include intersection scenarios and some pedestrian data; however, high-risk pedestrian–vehicle conflicts remain rare. Large multimodal datasets such as Waymo Open Motion (Sun et al. 2020), nuScenes (Caesar et al. 2020), Lyft Level 5 (Houston et al. 2021) extend coverage to urban environments using multimodal sensors. Although they include pedestrian and vehicle trajectories, most samples reflect routine traffic rather than safety-critical or evasive interactions (Rempe *et al.* 2022, Alozi and Hussein 2024). Drone-based datasets like rounD (Krajewski *et al.* 2020) and the HDI (Pu *et al.* 2025b) capture many dense intersection scenes but still lack naturalistic conflict cases.

Simulation-based datasets expand the range of scenarios but often fail to reproduce realistic behaviors. Virtual KITTI (Gaidon *et al.* 2016) extends the KITTI benchmark into a synthetic domain using Blender, offering structured urban trajectories but with scripted agents that lack realistic anticipation and responsiveness. Similarly, SYNTHIA (Ros *et al.* 2016), built on Unreal Engine, includes crosswalk scenarios but still shows behaviors that differ from real human decision-making. Game-based datasets like GTA V (Ros *et al.* 2016) provide visually rich pedestrian scenes from open-world simulations, but their artificial rules and arbitrary behaviors lead to unrealistic movements. Synscapes (Wrenninge and Unger 2018) offers photorealistic rendering and dense semantic labels, yet its agents remain non-interactive and lack human-like decisions. ApolloScape (Huang *et al.* 2018) , which rebuilds urban environments from real video, is limited by motion errors and unrealistic driving behaviors.

In summary, real-world datasets excel in realism but lack coverage of rare safety-critical scenarios, while simulation-generated datasets offer scenario diversity but often fail to capture authentic behaviors in naturalistic interactions (Grislain *et al.* 2024). These gaps show the need for hybrid methods that mix real-world behavior with scalable simulation. This can create large datasets of realistic near-miss events that are both credible and diverse enough to test ADS decision-making.

**Table 1 Summary of Real-World and Simulation-Generated Vehicle-Pedestrian Datasets**

| Data Source Type | Dataset | Data Collection Approach | Pedestrians and Vehicles Included | Driving Scenarios | Strengths | Limitations |
|---|---|---|---|---|---|---|
| Real-world | NGSIM (Alexiadis et al. 2004) | Fixed ground cameras | No | Highways | Widely used; rich vehicle data | No pedestrians; outdated scenes |
| | HighD (Krajewski et al. 2018) | UAV aerial videos | No | Highways | High accuracy vehicle tracking | limited interactions |
| | INTERACTION (Zhan et al. 2019) | UAV and fixed cameras | Partial | Roundabouts, urban intersections | Captures interactive behavior | Sparse coverage of pedestrian interactions |
| | inD (Bock et al. 2020) | UAV-based intersection recordings | Yes | Urban intersections | High-fidelity pedestrian and vehicle trajectories | Sparse coverage of pedestrian interactions |
| | rounD (Krajewski *et al.* 2020) | UAV over roundabouts | No | Roundabouts | Detailed vehicle merging dynamics | Limited pedestrian–vehicle interactions |
| | Waymo Open Motion (Sun et al. 2020) | Onboard LiDAR, camera | Yes | Urban, suburban roads | Multimodal dynamic data | Sparse safety-critical interactions |
| | nuScenes (Caesar et al. 2020) | Multisensor urban recordings | Yes | Urban | Comprehensive perception labels | Few safety-critical interactions |
| | Lyft Level 5 (Houston et al. 2021) | Onboard LiDAR, camera | Yes | Urban | Large scale; diverse conditions | Limited coverage of safety-critical events |
| | HDI (Pu *et al.* 2025b) | UAV recordings | Yes | Urban intersections | Dense vehicle-pedestrian flows | Sparse annotations of near-miss events |
| Simulation-Generated | Virtual KITTI (Gaidon *et al.* 2016) | Blender-based synthetic platform | Yes | Urban | Labeled vehicle-pedestrian tracks | Lacks naturalistic interactions |

| Data Source Type | Dataset | Data Collection Approach | Pedestrians and Vehicles Included | Driving Scenarios | Strengths | Limitations |
|---|---|---|---|---|---|---|
| | SYNTHIA (Ros *et al.* 2016) | Unreal Engine-based synthetic virtual city | Yes | Urban | Diverse crosswalk behavior | Agent behaviors not human-like |
| | GTA V-Based Datasets (Richter *et al.* 2016) | Extracted via GTA V game engine | Yes | Urban | Rich pedestrian interactions | Unrealistic traffic rules and behaviors |
| | Synscapes (Wrenninge and Unger 2018) | Proprietary synthetic rendering engine | Yes | Urban and suburban roads | Dense agent annotations | Agents lack human-like decision-making |
| | ApolloScape (Huang *et al.* 2018) | Synthetic 3D from real videos | Yes | Urban | High-resolution multimodal data | Agent motion lacks realism |

## 2.2. Reinforcement Learning for Behavior Modeling

Ensuring the safety of ADS in urban environments requires decision-making models capable of adapting to highly dynamic vehicle–pedestrian interactions (Bautista-Montesano *et al.* 2022). Early AV safety evaluations relied on rule-based approaches for lane changing, merging, and car-following (Wang *et al.* 2024, Yao and Sun 2025). While effective in structured scenarios, these methods could not generalize to uncertain or interactive environments (Tian *et al.* 2024, Ji *et al.* 2025). Reinforcement learning has emerged as a promising paradigm, enabling agents to develop adaptive strategies through iterative interactions with their environment (Wang *et al.* 2022, Yan *et al.* 2023). Reinforcement learning for autonomous driving is typically implemented in a single-stage manner, using either real-world datasets or synthetic simulation data (Hu *et al.* 2025).

Data-driven approaches train models on naturalistic real-world datasets, whereas simulation-driven approaches rely on synthetic environments. Models trained on real-world data, such as Waymo Open Motion (Sun et al. 2020) and Lyft Level 5 (Houston et al. 2021), reproduce naturalistic behavior but are limited by the scarcity of safety-critical events (Apostolovski et al. 2022). Conversely, simulation-based datasets such as CARLA (Dosovitskiy et al. 2017) and SUMO (Krajzewicz 2010) provide diverse risk scenarios but suffer from a reality gap where agent decisions diverge from human behavior (Dieter et al. 2023, Dave 2024). Within this single-stage paradigm, deep reinforcement learning (DRL) algorithms have been widely applied. For example, Deep Deterministic Policy Gradient (DDPG) (Lillicrap 2015) and its extensions demonstrate efficacy in trajectory planning, car-following, and collision avoidance (Tselentis and Papadimitriou 2023, Azfar *et al.* 2024). Similarly, PPO (Schulman *et al.* 2017) and SAC (Haarnoja *et* *al.* 2018) have been employed for continuous control tasks in simulation. However, these single-agent methods optimize only individual vehicle actions, neglecting the interactive responses of other agents such as pedestrians (Guo *et al.* 2023, Pu *et al.* 2026a). Consequently, multi-agent reinforcement learning, including MA-DDPG (Lowe *et al.* 2017), has been introduced to capture bidirectional feedback between vehicles and pedestrians (Zheng and Liu 2019, Zhang *et al.* 2023). Despite these advances, both data-driven and simulation-driven reinforcement learning approaches remain limited in scope. Approaches trained solely on real-world datasets learn naturalistic patterns yet fail to exploit simulation to validate or extend model generalizability (Almutairi *et al.* 2025). Conversely, approaches trained exclusively in simulation can generate diverse scenarios but fail to capture the authentic avoidance behaviors of real vehicles and pedestrians.

Beyond RL-based frameworks, two alternative paradigms have been explored for vehicle-pedestrian conflict simulation: generative models (Gupta et al. 2022) and inverse reinforcement learning (IRL) (Nasernejad et al. 2023). Generative models learn the underlying trajectory distribution from data without manually designed reward functions (Yin et al. 2021). TrafficSim (Suo et al. 2021) simulates realistic multi-agent behaviors by learning a generative prior over the joint distribution of vehicle interactions, while Safe-Sim (Chang et al. 2024) extends this to safety-critical scenario generation through diffusion-controllable adversarial agents. Although generative models flexibly capture multimodal behavioral distributions, they lack explicit safety constraints and online adaptation to novel environments (Zhou et al. 2025). IRL offers a complementary approach by recovering reward functions directly from expert demonstrations, bypassing manually specified reward design (Luo et al. 2025). Alozi and Hussein (2024) recovered context-dependent utility functions capturing pedestrian risk perception and social norms around autonomous vehicles. Lanzaro and Sayed (2025) applied a Markov game-based IRL framework to recover driver and pedestrian behavioral objectives across different interaction environments. Khuzam et al. (2025) further extended this approach to model how jaywalking alters pedestrian interaction strategies through multiagent utility inference. However, both generative models and IRL frameworks face generalization challenges in safety-critical scenarios underrepresented in naturalistic datasets, and neither supports continuous online adaptation.

Hybrid learning frameworks have been proposed to mitigate these shortcomings by combining real-world grounding with simulation-driven diversity. These methods typically pre-train policies on naturalistic datasets before deployment in simulation for performance validation and enhancement (Khan *et al.* 2018). Representative examples include BITS (Xu et al. 2023), which learns driver intent from human trajectories before applying it in synthetic tasks, AADS (Li *et al.* 2019), which fuses real images with simulated agents to improve visual realism, and RoboTron-Sim (Xiao *et al.* 2025), which generates edge cases such as sudden pedestrian crossings or cut-ins (Pu *et al.* 2026c). Other studies such as RALAD (Zuo *et al.* 2025) use retrieval-augmented domain adaptation to align simulated and real-world scenarios, while digital twin systems replicate intersections with high fidelity (Hu *et al.* 2023, Kamal *et al.* 2024). Although these hybrid learning approaches reduce the reality gap and enrich scenario diversity, they primarily validate first-stage models without supporting continuous adaptive learning in simulation. As a result, the generated data and policy behavior remain similar to first-stage outcomes, lacking transferability, mutual adaptation, and diversity in safety-critical interactions. This limitation keeps the agents "frozen" with the knowledge learned from the initial training data. As a result, their decision-making models are fragile and cannot adapt to new situations in dynamic simulation environments.

To bridge these gaps, there is a need for an advanced hybrid learning framework within a three-stage architecture that not only grounds agent policies in real-world data but also allows them to continuously adapt through the synergy of data-driven and simulation-driven learning.

## 3. Methodology

The three-stage framework is based on MA-SST-DDPG to systematically model realistic vehicle-pedestrian collision avoidance behaviors. This approach merges real-world data authenticity with simulation scalability to overcome key limitations in existing datasets. The framework operates in three sequential stages as illustrated in Figure 1.

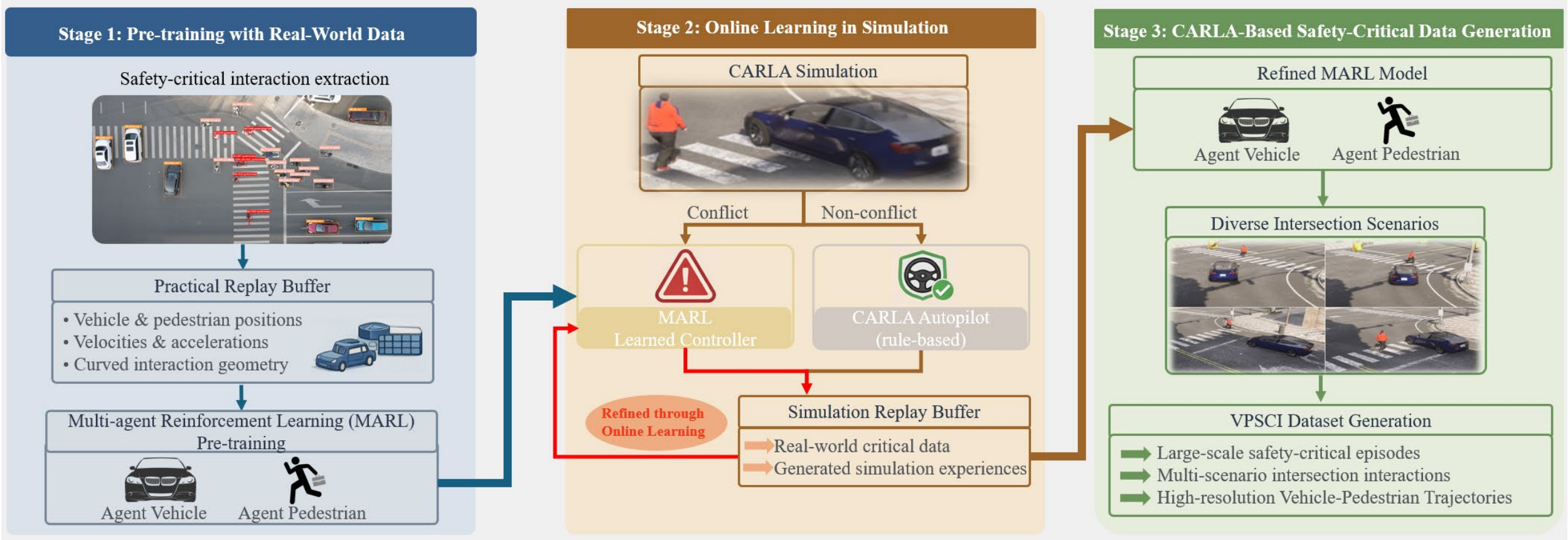

**Figure 1: Flowchart of three-stage framework for safety-critical data generation**

### Stage 1: Pre-training with Real-World Data

The process begins by extracting real-world safety-critical interactions from the HDI dataset using a CurvTTC threshold of 5 seconds. Unlike traditional TTC, CurvTTC (Pu et al. 2025b) provides a more realistic representation of intersection interactions by accounting for two-dimensional curved paths, capturing the non-linear trajectories characteristic of right-turn vehicle-pedestrian encounters. The 5-second threshold was specifically selected to capture the complete behavioral response window rather than just the moment of imminent impact. By using this wider window, the study can model the full interaction process, starting from the point where road users first perceive a potential conflict and initiate adjustments, through the entire evasive sequence, until the situation is resolved. Post-Encroachment Time (PET) (Minderhoud and Bovy 2001) is a common safety metric but it is unsuitable for highly dynamic interactions. PET can only be computed after both agents sequentially pass through a fixed conflict point. However, pedestrian behavior in safety-critical scenarios is highly variable. A pedestrian may perceive a threat and retreat to the curb instead of completing the crossing. In this case, the trajectories do not intersect and PET becomes undefined despite the extreme risk. Furthermore, PET assumes a fixed conflict point geometry, unsuitable for the curvilinear trajectories of right-turn interactions at complex urban intersections. CurvTTC addresses both limitations by providing a continuous conflict severity estimate across all interaction phases.

This provides a more realistic representation of their interactions. The dataset is available at: https://github.com/Qpu523/HDI-Dataset. These interactions are characterized by detailed kinematic parameters, including lateral and longitudinal speeds $\left(v_{veh}^{x}, v_{veh}^{y}, v_{ped}^{x}, v_{ped}^{y}\right)$, distances ( $d_{vp}^{x}, d_{vp}^{y}, d_{pv}^{x}, d_{pv}^{y}$ ), and relative motion ( $\Delta v_{vp}^{x}, \Delta v_{vp}^{y}, \Delta v_{pv}^{x}, \Delta v_{pv}^{y}$ ). This curated information is stored in a Practical Replay Buffer as the learning foundation.

This real-world data is utilized to pre-train the dual-agent networks for vehicles and pedestrians within the MA-SST-DDPG framework. During this phase, the SST Actor networks receive historical observation sequences ( $O_{veh}(t), O_{ped}(t)$ ) and employ Transformer-based state-space modeling to generate two-dimensional acceleration actions $\left(a_{veh}^{x}(t), a_{veh}^{y}(t), a_{ped}^{x}(t), a_{ped}^{y}(t)\right)$. Concurrently, the SST Critic evaluates these actions through state-action pairs, outputting Q-values to guide gradient-based policy updates toward realistic evasive behaviors.

**Stage 2: Online Learning in a Simulation Environment**

The pre-trained MA-SST-DDPG agents are deployed in CARLA's Town10 map, which features diverse intersection layouts, randomized spawn locations, and varied motion patterns. A monitoring module continuously evaluates CurvTTC. When CurvTTC < 5 s, the interaction is identified as safety-critical, triggering the intervention mechanism. For non-conflict interactions where CurvTTC ≥ 5 s, control remains with the rule-based CARLA autopilot without learning intervention. Upon activation, the CARLA autopilot is overridden, and the MA-SST-DDPG framework assumes control to generate evasive maneuvers via learned acceleration outputs. During these events, the corresponding state–action–reward tuples data are stored in a simulation replay buffer. The model iteratively updates its policy through online learning driven by safety-critical intervention events, using the simulation replay buffer to refine evasive control behaviors. This approach improves adaptability across intersection types and enhances generalization to complex urban environments, thereby improving behavioral consistency within simulation environments. As a result, this process yields the Refined MA-SST-DDPG model, demonstrating enhanced generalization across diverse traffic scenarios.

**Stage 3: CARLA-Based Large-Scale Data Generation**

In Stage 3, the Refined MA-SST-DDPG model is used to generate large-scale safety-critical interactions within the CARLA simulator. The agents operate in multiple intersection scenarios that mimic urban traffic. The simulation uses random initial values for vehicle speed, pedestrian speed, and positions to create diverse episodes. When CurvTTC < 5 s, the situation is regarded as a safety-critical scenario, and the learned policy is activated to produce evasive maneuvers. This process ensures each episode shows realistic evasive actions under high-risk conditions. CARLA's physics engine updates all agent states, including position, speed, and acceleration, at every step. The system saves all trajectories and interaction data in structured files to build the VPSCI dataset, which comprises large-scale safety-critical episodes, multi-scenario intersection interactions, and high-resolution vehicle–pedestrian trajectories.

The behavioral realism of interactions in the VPSCI dataset was validated through Turing test-based human evaluation. Participants compared real trajectories with those generated by the three-stage framework and CARLA baselines. Furthermore, the dataset's authenticity was validated through comparison with real-world datasets (Argoverse 2 and TGSIM). A safety analysis further examines conflict rates and pedestrian yielding behaviors. The detailed three-stage framework is presented in Algorithm 1.

Algorithm 1: Three-Stage Framework Based on MA-SST-DDPG

**Initialization**

Randomly initialize SST critic networks: $Q_{Ag}\left(H_{Ag}(t), A_{Ag}(t) \mid \theta_{Q_{Ag}}\right)|_{Ag=veh,ped}$ with weights $\theta_{Q_{veh}}$ and $\theta_{Q_{ped}}$.

Initialize SST actor networks: $\mu_{Ag}\left(H_{Ag}(t) \mid \theta_{\mu_{Ag}}\right)$ with weights $\theta_{\mu_{veh}}$ and $\theta_{\mu_{ped}}$.

Initialize SST target networks $Q'_{veh}, Q'_{ped}, \mu'_{veh}$, and $\mu'_{ped}$ with corresponding weights:

$$\theta_{Q'_{veh}} \leftarrow \theta_{Q_{veh}}, \theta_{Q'_{ped}} \leftarrow \theta_{Q_{ped}};$$
$$\theta_{\mu'_{veh}} \leftarrow \theta_{\mu_{veh}}, \theta_{\mu'_{ped}} \leftarrow \theta_{\mu_{ped}}.$$

Initialize two separate replay buffers:

Practical Replay Buffer ( $B_p$ ): For Stage 1, using real-world safety-critical data.

Simulation Replay Buffer ( $B_s$ ): For Stage 2, using CARLA-based interaction data.

Initialize a random process: $\mathcal{N}$ for action exploration.

**Stage 1: Pre-training with Real-World Data**

Objective: To learn realistic, human-like evasive behaviors from actual safety-critical data.

Fill the Practical Replay Buffer ( $B_p$ ) with safety-critical interaction sequences $\{O_{Ag}(1), A_{Ag}(1), R_{Ag}(1), \ldots, O_{Ag}(T), A_{Ag}(T), R_{Ag}(T)\}$ extracted from the HDI dataset.

For each training iteration do:

Sample a random mini-batch of $M$ samples from $B_p$.

For each sample $k$ in the mini-batch, construct the history:

$$H_{Ag}^k(t) = \left(O_{Ag}^k(1), A_{Ag}^k(1), \ldots, O_{Ag}^k(t-1), A_{Ag}^k(t-1), O_{Ag}^k(t)\right).$$

Calculate target values:

$$y_{\text{veh}}^k(t) = R_{\text{veh}}^k(t) + \gamma Q'_{\text{veh}}\left(H_{\text{veh}}^k(t+1), A_{\text{veh}}^k(t+1), A_{\text{ped}}^k(t+1)\right)\Big|_{A_{\text{veh}}^k(t+1)=\mu'_{\text{veh}}\left(H_{\text{veh}}^k(t+1)\right)}$$

$$y_{\text{ped}}^k(t) = R_{\text{ped}}^k(t) + \gamma Q'_{\text{ped}}\left(H_{\text{ped}}^k(t+1), A_{\text{ped}}^k(t+1), A_{\text{veh}}^k(t+1)\right)\Big|_{A_{\text{ped}}^k(t+1)=\mu'_{\text{ped}}\left(H_{\text{ped}}^k(t+1)\right)}$$

Update critic network:

$$L\left(\theta_{Q_{\text{veh}}}\right) = \frac{1}{M}\sum_k \sum_t \left(y_{\text{veh}}^k(t) - Q_{\text{veh}}\left(H_{\text{veh}}^k(t), A_{\text{veh}}^k(t), A_{\text{ped}}^k(t)\right)\right)^2$$
$$L\left(\theta_{Q_{\text{ped}}}\right) = \frac{1}{M}\sum_k \sum_t \left(y_{\text{ped}}^k(t) - Q_{\text{ped}}\left(H_{\text{ped}}^k(t), A_{\text{veh}}^k(t), A_{\text{ped}}^k(t)\right)\right)^2$$

Update SST actor networks using policy gradient:

$$\nabla_{\theta_{\mu_{\text{veh}}}} J \approx \frac{1}{M}\sum_{k}\sum_{t} \nabla_{\theta_{\mu_{\text{veh}}}} \mu_{\text{veh}}\left(H_{\text{veh}}^{k}(t)\right) \nabla_{A_{\text{veh}}^{k}(t)} Q_{\text{veh}}\left(H_{\text{veh}}^{k}(t), A_{\text{veh}}^{k}(t), A_{\text{ped}}^{k}(t)\right)\Big|_{A_{\text{veh}}^{k}(t)=\mu_{\text{veh}}\left(H_{\text{veh}}^{k}(t)\right)}$$

$$\nabla_{\theta_{\mu_{\text{ped}}}} J \approx \frac{1}{M}\sum_{k}\sum_{t} \nabla_{\theta_{\mu_{\text{ped}}}} \mu_{\text{ped}}\left(H_{\text{ped}}^{k}(t)\right) \nabla_{A_{\text{ped}}^{k}(t)} Q_{\text{ped}}\left(H_{\text{ped}}^{k}(t), A_{\text{veh}}^{k}(t), A_{\text{ped}}^{k}(t)\right)\Big|_{A_{\text{ped}}^{k}(t)=\mu_{\text{ped}}\left(H_{\text{ped}}^{k}(t)\right)}$$

Soft update SST target networks:

$$\theta_{Q'_{Ag}} \leftarrow \tau\theta_{Q_{Ag}} + (1-\tau)\theta_{Q'_{Ag}},\ \theta_{\mu'_{Ag}} \leftarrow \tau\theta_{\mu_{Ag}} + (1-\tau)\theta_{\mu'_{Ag}} |_{A=veh,ped}$$

end for

**Stage 2: Online Learning in a Simulation Environment**

Objective: To refine the pre-trained agents for a wide variety of high-risk scenarios within the CARLA simulator.

For each episode (from 1 to M) in CARLA do:

Initialize agent histories $H_{\text{veh}}(0)$ and $H_{\text{ped}}(0)$.

For each time step $t$ (from 1 to T ) do:

Receive current observations: $O_{veh}(t)$ and $O_{ped}(t)$.

Safety Check: If CurvTTC < 5 s, it activates the MA-SST-DDPG model to take control.

If the MA-SST-DDPG model is active:

Construct histories: $H_{Ag}(t) \leftarrow \left(H_{Ag}(t-1), A_{Ag}(t-1), O_{Ag}(t)\right)$.

Select actions: $A_{Ag}(t) = \mu_{Ag}\left(H_{Ag}(t) \mid \theta_{\mu_{Ag}}\right) + \mathcal{N}$.

Store the sequence in the Simulation Replay Buffer ( $B_s$ ).

End For

Online Training:

Sample a random mini-batch of $M$ samples from $B_s$.

Perform all network updates (calculate targets, update critics, update actors, and soft update targets) using the same formulas as in Stage 1.

End For

**Stage 3: CARLA-Based Large-Scale Data Generation**

Objective: To utilize the refined model to generate the VPSCI dataset.

Deploy the Refined MA-SST-DDPG agents in CARLA across eight distinct intersection scenarios.

## 3.1. Pre-training with Real-World Data

The core of this framework is the MA-SST-DDPG model (Pu *et al.* 2026b), designed to capture the temporal dependencies and critical features of safety-critical vehicle-pedestrian interactions. Built upon the MA-DDPG architecture (Lowe et al. 2017), it incorporates state-space modeling (Gu and Dao 2023) and Transformer-based mechanisms (Han et al. 2022) through a State-Space Transformer (SST) module.

The interaction geometry underlying the observation space is illustrated in Figure 2. The vehicle velocity $v_{veh}$ is decomposed into a longitudinal component $v_{veh}^{y}$ along the vehicle's heading direction and a lateral component $v_{veh}^{x}$ perpendicular to it. The pedestrian velocity $v_{ped}$ is defined along the crossing direction, and the red circle denotes the projected conflict point where the vehicle and pedestrian trajectories are anticipated to intersect. This coordinate system is anchored to the vehicle's instantaneous heading at each time step, ensuring geometric consistency under curved approach paths and arbitrary crossing angles characteristic of urban right-turn scenarios.

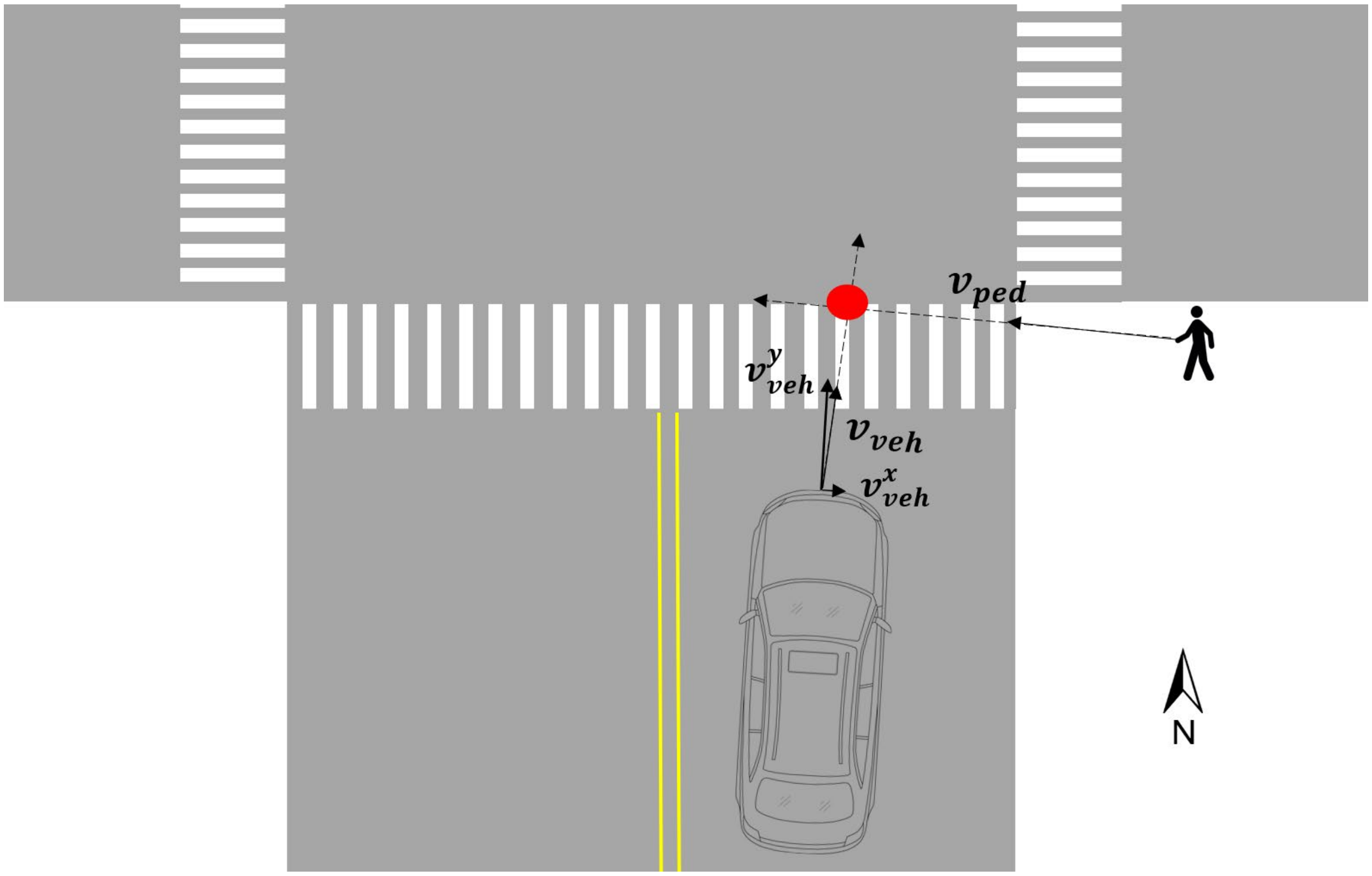

**Figure 2: Illustration of vehicle-pedestrian interaction geometry at conflict onset**

Vehicle-pedestrian safety-critical interactions are modeled as a two-agent partially observable Markov game, defined as $\langle \mathcal{N}, \{\mathcal{O}_{Ag}\}, \{\mathcal{A}_{Ag}\}, \mathcal{T}, \{R_{Ag}\}, \gamma \rangle$, where $\mathcal{N} = \{veh, ped\}$, $\mathcal{O}_{Ag}$ and $\mathcal{A}_{Ag}$ denote the local observation and action spaces of agent $Ag, \mathcal{T}$ is the state transition function, and $\gamma$ is the discount factor. Each agent receives only its local observation rather than the full global state, reflecting the realistic condition that neither road user has complete knowledge of the other's intentions. The game is formulated as a non-zero-sum cooperative rather than a zero-sum competitive game. In a zero-sum formulation, one agent's reward gain constitutes the other's loss, incentivizing adversarial behavior that is incompatible with collision avoidance, where both agents share a mutual interest in avoiding collision. The equilibrium concept is a Nash equilibrium in decentralized stochastic policies, where neither agent can improve its expected cumulative reward by unilaterally deviating from its policy given the other agent's policy:

$$V_{veh}(\pi_{veh}^*, \pi_{ped}^*) \geq V_{veh}(\pi_{veh}, \pi_{ped}^*), \qquad \forall \pi_{veh} \tag{1}$$

$$V_{ped}(\pi_{veh}^*, \pi_{ped}^*) \geq V_{ped}(\pi_{veh}^*, \pi_{ped}), \qquad \forall \pi_{ped} \tag{2}$$

This equilibrium is approximated through the centralized training with decentralized execution paradigm. During training, each agent's critic receives the joint action tuple $(A_{veh}, A_{ped})$, enabling Q-value estimation conditioned on both agents' behaviors and progressive convergence toward a mutual best-response configuration. During execution, each actor operates solely on its local observations.

The SST module is composed of three components that work together to enhance temporal modeling and prioritize safety-critical scenarios. Sequential observations $x_t \in \mathbb{R}^d$ are processed through structured state transitions parameterized by learnable matrices $A, B, C, D$, where the hidden state $h_t \in \mathbb{R}^d$ evolves according to

$$h_t = Ah_{t-1} + Bx_t, y_t = Ch_t + Dx_t \tag{3}$$

Here, $h_t$ represents the latent dynamics of the system and $y_t$ denotes the output at time $t$. This formulation captures long-term temporal dependencies with linear computational complexity $\mathcal{O}(L)$, in contrast to the quadratic complexity $\mathcal{O}(L^2)$ of standard Transformers. A selective attention mechanism then applies dynamic gating based on input relevance, with attention weights computed as

$$\alpha_t = \text{softmax}\left(\frac{W_q h_t \cdot W_k h_{1:t}^\top}{\sqrt{d_k}}\right) \tag{4}$$

where $W_q, W_k \in \mathbb{R}^{d \times d_k}$ are learnable projection matrices and $d_k$ is the key dimension. When CurvTTC $<$ $5s$, this mechanism prioritizes features contributing to imminent risk. Finally, importance weighting is applied to emphasize high-impact samples by computing

$$\omega_i = \exp\left(\frac{|\ \text{TD_error}_i\ |}{\tau}\right) \tag{5}$$

where $\text{TD_error}_i$ is the temporal-difference error of sample $i$ and $\tau > 0$ is a temperature parameter that controls the strength of reweighting. This ensures that near-miss events receive greater emphasis during training, improving the model's ability to capture sparse but critical safety scenarios.

As shown in Figure 3, the architecture includes two agents: vehicle (veh) and pedestrian (ped), each with four networks: SST actor ( $\mu_{\text{veh}}$ or $\mu_{\text{ped}}$ ), SST target actor ( $\mu'_{\text{veh}}$ or $\mu'_{\text{ped}}$ ), SST critic ( $Q_{\text{veh}}$ or $Q_{\text{ped}}$ ), and SST target critic ( $Q'_{\text{veh}}$ or $Q'_{\text{ped}}$ ). The actor networks $(\mu)$ interact with the environment to generate actions ( $A^k_{\text{veh}}, A^k_{\text{ped}}$ ) based on observed states ( $O^k_{\text{veh}}, O^k_{\text{ped}}$ ), while the critic networks $(Q)$ evaluate these actions by producing Q-values that guide policy updates. For stable learning, the target networks ( $\mu'$ and $Q'$ ) are softly updated as delayed versions of the actor and critic networks. The SST module reweights data samples using importance weights ($\omega_{veh}$, $\omega_{ped}$) and temporal difference errors ($\delta_{veh}$, $\delta_{ped}$) to construct mini-batches for training, enabling the model to adapt to safety-critical behaviors by emphasizing high-impact interactions and capturing complex temporal dependencies. The main notations used in this study are summarized in Table 2.

**Table 2 Summary of notations**

| Symbol | Definition |
|---|---|
| Ag | Ag ∈ {veh, ped}, where Ag denotes the agent (vehicle or pedestrian) |
| $t, T$ | Time, incrementing by 0.1 s |
| $\Delta t$ | Time interval |
| $v_{veh}^{y}(t), v_{ped}^{y}(t)$ | Longitudinal speed of the vehicle and pedestrian at time $t$ |
| $v_{veh}^{x}(t), v_{ped}^{x}(t)$ | Lateral speed of the vehicle and pedestrian |
| $d_{vp}^{y}(t), d_{pv}^{y}(t)$ | Longitudinal distance between the vehicle and pedestrian, and vice versa |
| $d_{vp}^{x}(t), d_{pv}^{x}(t)$ | Lateral distance between the vehicle and pedestrian, and vice versa |
| $\Delta v_{vp}^{y}(t), \Delta v_{pv}^{y}(t)$ | Longitudinal relative speed between the vehicle and pedestrian, and vice versa |
| $\Delta v_{vp}^{x}(t), \Delta v_{pv}^{x}(t)$ | Lateral relative speed between the vehicle and pedestrian, and vice versa |
| $a_{veh}^{y}(t), a_{ped}^{y}(t)$ | Longitudinal acceleration of the vehicle and pedestrian |
| $a_{veh}^{x}(t), a_{ped}^{x}(t)$ | Lateral acceleration of the vehicle and pedestrian |
| $O_{veh}(t), O_{ped}(t)$ | Observation of the vehicle and pedestrian |
| $A_{veh}(t), A_{ped}(t)$ | Action of the vehicle and pedestrian |
| $H_{veh}(t), H_{ped}(t)$ | Observation-action history of vehicle and pedestrian |
| $R_{veh}(t), R_{ped}(t)$ | Reward of the vehicle and pedestrian |

In this framework, vehicles and pedestrians act as agents that continuously observe and respond to each other. Their actions are defined in continuous spaces by longitudinal and lateral accelerations. Agent policies are pre-trained using real-world safety-critical interactions from the HDI dataset.

At each time step $t$, each agent receives its local observation. Specifically, the vehicle receives an observation $O_{veh}(t)$ that includes its own kinematic information and the relative distance and speed to the pedestrian. The observation of the vehicle at the time $t$ is given by:

$$O_{veh}(t) = \left(v_{veh}^{y}(t), v_{veh}^{x}(t), d_{vp}^{y}(t), d_{vp}^{x}(t), \Delta v_{vp}^{y}(t), \Delta v_{vp}^{x}(t)\right) \tag{6}$$

where the longitudinal and lateral distances are computed directly from the position coordinates of the two agents: $d_{vp}^{y}(t) = y_{veh}(t) - y_{ped}(t)$ and $d_{vp}^{x}(t) = x_{veh}(t) - x_{ped}(t)$ . The pedestrian $j$ receives an observation $O_j(t)$, which includes its kinematic information and the relative distance and speed of the vehicle $i$ :

$$O_{ped}(t) = \left(v_{ped}^{y}(t), v_{ped}^{x}(t), d_{pv}^{y}(t), d_{pv}^{x}(t), \Delta v_{pv}^{y}(t), \Delta v_{pv}^{x}(t)\right) \tag{7}$$

Pedestrian heading is approximated by the direction of the instantaneous velocity vector $\left(v_{ped}^{x}, v_{ped}^{y}\right)$ derived from the trajectory coordinates, since body orientation cannot be reliably extracted from bird's-eye UAV video. The actions of the vehicle and pedestrian are defined as $A_{Ag}(t) = \left(\hat{a}_{Ag}^{y}(t), \hat{a}_{Ag}^{x}(t)\right)\Big|_{Ag=veh,ped}$, where $\hat{a}^{y}(t)$ and $\hat{a}^{x}(t)$ represent the longitudinal and lateral accelerations for both agents, respectively. Through these two-dimensional accelerations, directional adjustments of pedestrian movement, such as lateral avoidance and path deviation, are implicitly represented as changes in the velocity direction. Each action represents a singlestep prediction, and the full trajectory is reconstructed by sequentially applying the policy at each time step.

After receiving the observation $O_{Ag}(t)|_{Ag=veh,ped}$, each agent (vehicle and pedestrian) constructs a history $H_{Ag}(t) = \left(O_{Ag}(1), A_{Ag}(1), \ldots, O_{Ag}(t-1), A_{Ag}(t-1), O_{Ag}(t)\right)|_{Ag=veh,ped}$ by combining past observations and actions with the current observation. The agent then selects an action based on this history and its policy: $A_{Ag}(t) = \mu_{Ag}\left(H_{Ag}(t) \mid \theta_{\mu_{Ag}}\right)|_{Ag=veh,ped}$.

The environment is then updated using the following kinematic model:

$$\hat{v}_{Ag}^{x}(t+1) = v_{Ag}^{x}(t) + \hat{a}_{Ag}^{x}(t)\Delta t|_{Ag=veh,ped} \tag{8}$$

$$\hat{v}_{Ag}^{y}(t+1) = v_{Ag}^{y}(t) + \hat{a}_{Ag}^{y}(t)\Delta t|_{Ag=veh,ped} \tag{9}$$

$$\Delta\hat{v}_{vp}^{y}(t+1) = \hat{v}_{veh}^{y}(t+1) - \hat{v}_{ped}^{y}(t+1) \tag{10}$$

$$\Delta\hat{v}_{vp}^{x}(t+1) = \hat{v}_{veh}^{x}(t+1) - \hat{v}_{ped}^{x}(t+1) \tag{11}$$

$$\Delta\hat{v}_{pv}^{y}(t+1) = \hat{v}_{ped}^{y}(t+1) - \hat{v}_{veh}^{y}(t+1) \tag{12}$$

$$\Delta\hat{v}_{pv}^{x}(t+1) = \hat{v}_{ped}^{x}(t+1) - \hat{v}_{veh}^{x}(t+1) \tag{13}$$

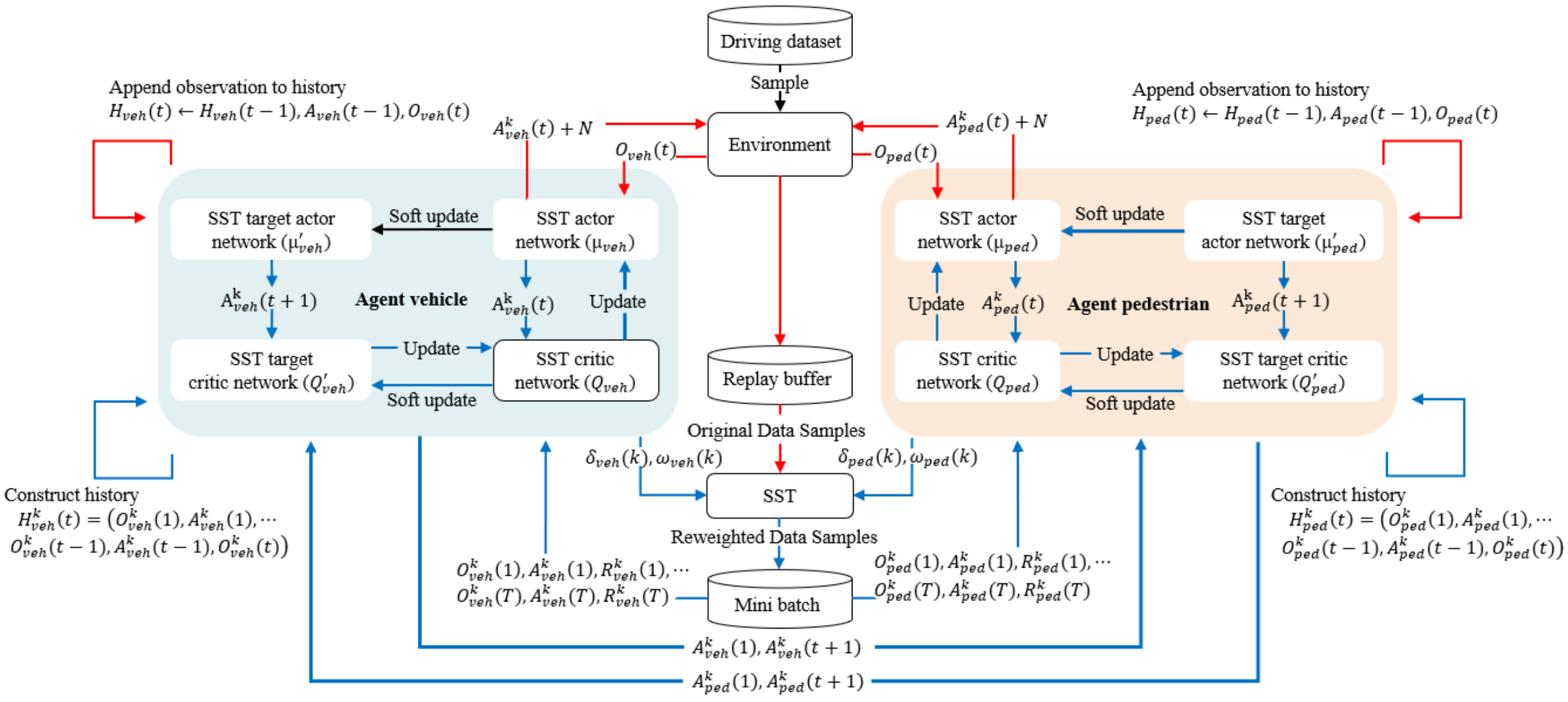

**Figure 3: Architecture of MA-SST-DDPG, including experience generation (red arrows) and training (blue arrows) procedure (Pu *et al.* 2026b)**

A speed reward function is introduced to penalize deviations from typical driving and walking speeds, thereby encouraging human-like behavior. The reward function is defined as:

$$R_{v_{veh}} = -\left[\left(\frac{\hat{v}^y_{veh}(t+1) - \mu_{v^y_{veh}}}{\sigma_{v^y_{veh}}} - \frac{v^y_{veh}(t+1) - \mu_{v^y_{veh}}}{\sigma_{v^y_{veh}}}\right)^2 + \left(\frac{\hat{v}^x_{veh}(t+1) - \mu_{v^x_{veh}}}{\sigma_{v^x_{veh}}} - \frac{v^x_{veh}(t+1) - \mu_{v^x_{veh}}}{\sigma_{v^x_{veh}}}\right)^2\right] \quad (14)$$

$$R_{v_{ped}} = -\left[\left(\frac{\hat{v}^y_{ped}(t+1) - \mu_{v^y_{ped}}}{\sigma_{v^y_{ped}}} - \frac{v^y_{ped}(t+1) - \mu_{v^y_{ped}}}{\sigma_{v^y_{ped}}}\right)^2 + \left(\frac{\hat{v}^x_{ped}(t+1) - \mu_{v^x_{ped}}}{\sigma_{v^x_{ped}}} - \frac{v^x_{ped}(t+1) - \mu_{v^x_{ped}}}{\sigma_{v^x_{ped}}}\right)^2\right] \quad (15)$$

where $R_{v_{veh}}$ is the speed reward for the vehicle, and $R_{v_{ped}}$ is the speed reward for the pedestrian. The terms $\mu_{v^y_{veh}}, \sigma_{v^y_{veh}}, \mu_{v^x_{veh}}, \sigma_{v^x_{veh}}, \mu_{v^y_{ped}}, \sigma_{v^y_{ped}}, \mu_{v^x_{ped}}$, and $\sigma_{v^x_{ped}}$ are the means and standard deviations of the longitudinal and lateral speeds for the vehicle and pedestrian, respectively. These normalized rewards guide the agents toward human-like motion while allowing adaptive responses to dynamic situations. Importantly, the reward design penalizes deviations between predicted and observed states rather than enforcing fixed statistical averages, enabling the policy to learn flexible collision avoidance strategies instead of rigidly reproducing site-specific behaviors.

For each agent, the mean ($\mu$) and standard deviation ($\sigma$) of longitudinal and lateral speeds are computed separately from the pre-training corpus prior to training and held fixed throughout Stage 1. Specifically, four statistical parameters are extracted $\mu_{v^y_{veh}}, \sigma_{v^y_{veh}}, \mu_{v^x_{veh}}, \sigma_{v^x_{veh}}$ for the vehicle, and $\mu_{v_{ped}^y}, \sigma_{v_{ped}^y}, \mu_{v_{ped}^x}, \sigma_{v_{ped}^x}$ for the pedestrian. The speed reward penalizes the normalized deviation between predicted and observed speeds in both longitudinal and lateral dimensions, reaching zero when the predicted speed exactly matches the observed speed and becoming increasingly negative as the discrepancy grows. The sum of longitudinal and lateral components ensures that both dimensions of motion must simultaneously align with observed speed profiles to minimize penalty. An identical formulation applies to the pedestrian agent using pedestrian-specific parameters.

To assess the representativeness of the 336 pre-training interactions, a sensitivity analysis was conducted by training the MA-SST-DDPG model under identical hyperparameter settings using subsets of 50, 100, 150, 200, 250, and 336 interactions, recording average episodic reward and critic loss at convergence for each configuration. Since the pre-training data exclusively targets safety-critical interactions (CurvTTC <5 s), a compact dataset suffices to capture the essential evasive response patterns.

**Table 3 Sensitivity analysis of pre-training sample size**

| Pre-training Samples | Vehicle Agent Reward | Pedestrian Agent Reward | Critic Loss |
|---|---|---|---|
| 50 | -6.83 | -7.21 | 0.4812 |
| 100 | -5.47 | -6.03 | 0.3654 |
| 150 | -4.12 | -4.89 | 0.2431 |
| 200 | -2.14 | -3.67 | 0.1124 |
| 250 | -2.09 | -3.61 | 0.1098 |
| 336 (full) | -1.94 | -3.52 | 0.1073 |

As shown in Table 3, both agent rewards and critic loss improve substantially as sample size increases from 50 to 200 interactions, with vehicle agent reward rising from −6.83 to −2.14 and critic loss declining from 0.4812 to 0.1124. Beyond 200 samples, all three metrics stabilize, with reward changes below 0.20 and critic loss changes below 0.006, indicating that 200 interactions suffice to capture the core kinematic patterns of safety-critical evasive behavior. The rapid convergence reflects the behavioral homogeneity of the pre-training data, which exclusively targets a single well-characterized behavioral regime (CurvTTC <5 s). These results confirm that the 336-sample dataset is both sufficient and representative of the target behavioral distribution.

### 3.2. Online Learning in Simulation

After pre-training, the agents are deployed in the CARLA (version 0.9.15) simulation environment to refine and validate their learned evasive strategies. CARLA, an open-source simulator built on Unreal Engine, adopts a client–server structure for autonomous driving research. The server renders the environment and simulates physical and kinematic processes. The Python client handles simulation logic, data collection, and agent control.

This study built a high-fidelity simulation in CARLA to analyze vehicle-pedestrian collision avoidance at intersections. The simulation uses Town10, a map with complex urban intersections, multiple traffic lanes, and pedestrian crosswalks. This setup supports realistic modeling of vehicle-pedestrian interactions in high-risk urban scenarios.

Figure 4 presents two representative intersections (Location A and Location B) selected for detailed analysis. These sites were chosen due to their frequent vehicle-pedestrian conflicts during right-turn movements, providing suitable conditions for examining safety-critical scenarios and evasive behaviors. Right-turn movements are largely unsignalized under North American and many international traffic regulations, allowing vehicles to turn continuously throughout the signal cycle and persistently share the crossing zone with pedestrians. Unlike left-turn and straight-through movements, which are temporally separated from pedestrian phases by signal control or oncoming traffic, right-turn conflicts are unmitigated, making them the highest-frequency vehicle-pedestrian interaction type at signalized urban intersections. Red arrows denote vehicle right-turn paths, and green arrows indicate pedestrian crossing directions.

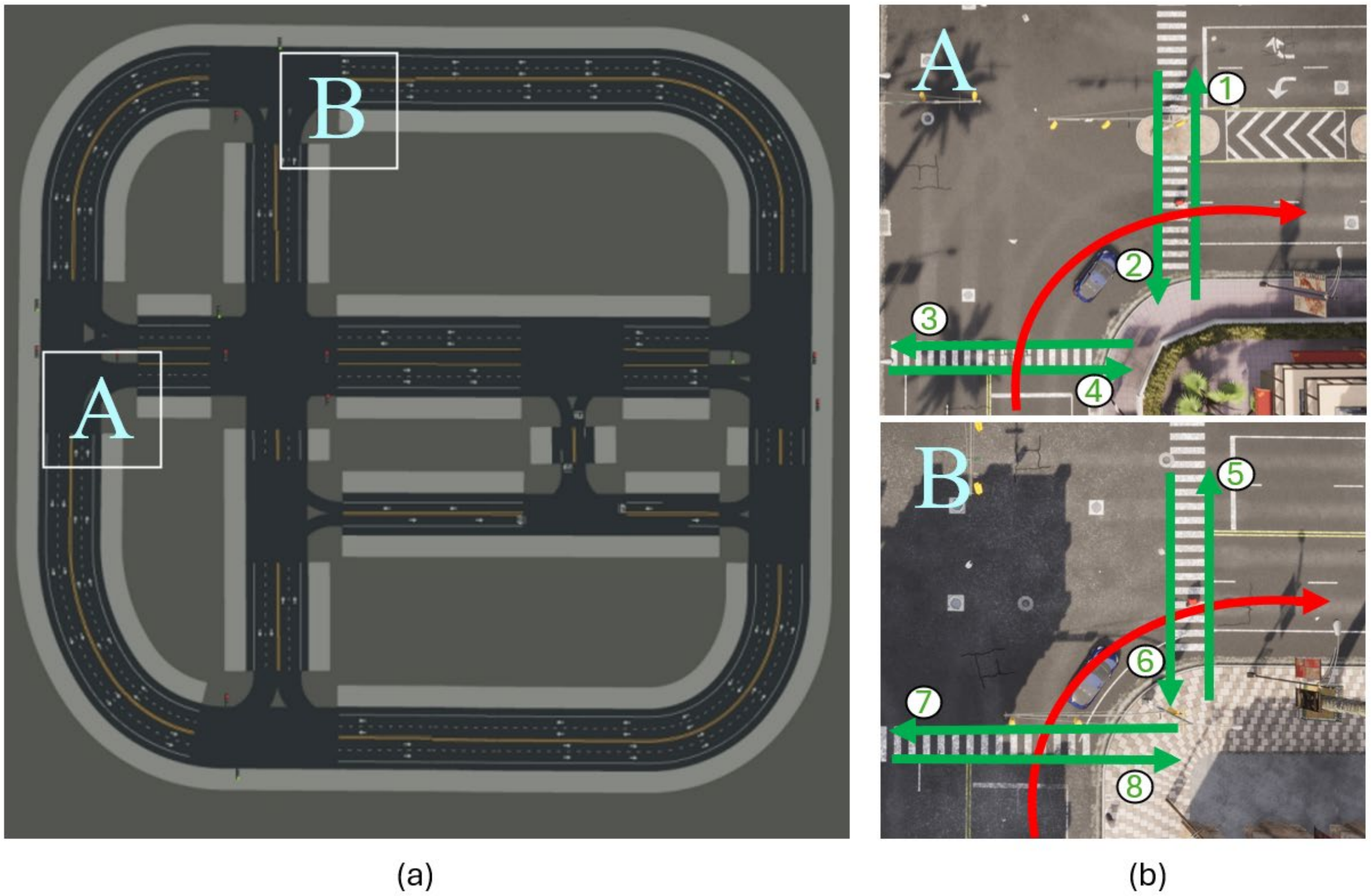

**Figure 4: Study area and interaction scenarios used in the simulation. (a) The layout of the CARLA simulation environment. (b) Detailed views of intersections A and B**

The simulation used two agent types: a Tesla Model 3 vehicle and a pedestrian model from the CARLA blueprint library. The vehicle operated in two modes. One mode used the built-in Traffic Manager for autonomous navigation. The other mode applied a MA-SST-DDPG model for adaptive control, which was activated when a collision risk (CurvTTC $<$ 5 s) was detected. The pedestrian walked with a stochastic pattern but adjusted movement when vehicles were nearby, using the same MA-SST-DDPG model.

Pedestrians were programmed to maintain forward progress toward the opposite side of the street, thereby maximizing the likelihood of interactions with turning vehicles.

To introduce variability, the simulation randomized agent starting positions, with vehicle throttle set between 0.6 and 0.9 and pedestrian speeds assigned randomly between 1 and 4 m/s. All simulations ran in synchronous mode with a fixed time step of 0.05 seconds to keep data generation and evaluation consistent.

A CurvTTC-based conflict check monitors vehicle–pedestrian states during simulation to prevent imminent collisions. CurvTTC considers curved and non-linear paths, offering a more accurate collision risk measure at intersections than traditional TTC (Pu *et al.* 2025a). When CurvTTC $< 5$ s, indicating high-risk condition, the MA-SST-DDPG policy overrode CARLA's default navigation and assumed control.

During the online learning process, the framework employs an adaptive exploration strategy that balances safety requirements with policy improvement. The exploration noise follows a time-dependent decay schedule:

$$\mathcal{N}(t) = \mathcal{N}_0 \cdot e^{-\beta t} + \mathcal{N}_{\min} \tag{16}$$

where $\mathcal{N}_0 = 0.3$ represents the initial exploration level, $\beta = 0.001$ controls the decay rate, and $\mathcal{N}_{\min} = 0.05$ ensures minimum exploration is maintained throughout training. This approach allows agents to explore new strategies early in training while gradually shifting toward exploitation of learned policies as performance stabilizes. The safety-critical nature of vehicle–pedestrian interactions requires a careful balance. Insufficient exploration can lead to poor collision-avoidance strategies. Excessive exploration can result in unsafe behaviors.

Beyond exploration management, preventing catastrophic forgetting of real-world behaviors learned in Stage 1 represents another critical challenge in online learning. First, the Simulation Replay Buffer maintains a portion of real-world experiences from the Practical Replay Buffer, ensuring continued exposure to authentic human behaviors during policy updates. Second, the reward function incorporates speed consistency terms that penalize deviations from realistic movement patterns observed in real-world data. Third, the learning rate for online adaptation is set lower than the initial pre-training phase ( $\alpha_{\text{online}} = 0.00005$ compared to $\alpha_{\text{pretrain}} = 0.0001$ ) to prevent rapid overwriting of learned behaviors. Additionally, experience replay sampling employs importance weighting to maintain representation of critical real-world scenarios:

$$P(i) = \frac{|\delta_i|^\alpha + \beta \cdot I_{\text{real}}(i)}{\sum_j \left(\left|\delta_j\right|^\alpha + \beta \cdot I_{\text{real}}(j)\right)} \tag{17}$$

where $\delta_i$ is the temporal difference error, $I_{\text{real}}(i)$ is an indicator function that equals 1 for real-world experiences and 0 for simulation experiences, and $\beta = 0.2$ controls the relative importance of maintaining real-world knowledge.

The model uses three learning mechanisms: adaptive exploration, knowledge preservation, and continuous policy refinement. These mechanisms work together to enable human-like agent behavior across diverse scenarios, with learned acceleration actions subsequently converted into control commands for both vehicle and pedestrian agents. For the vehicle agent, the SST-Actor network produces target lateral and longitudinal accelerations. These were converted into CARLA control commands using a dynamics-based process:

$$\text{steer} = \text{clamp}\left(\frac{\arctan\left(\frac{L \cdot a_{\text{veh}}^{x}}{{v_{veh}^{y}}^{2} + \epsilon}\right)}{\delta_{\max}}, -1.0, 1.0\right) \tag{18}$$

$$\text{throttle} = \begin{cases} \min\left(\frac{a_{\text{veh}}^{y}}{a_{\text{max_accel}}}, 1.0\right) & \text{if } a_{\text{veh}}^{y} \geq 0 \\ 0 & \text{otherwise} \end{cases} \tag{19}$$

$$\text{brake} = \begin{cases} \min\left(\frac{|a_{\text{veh}}^{y}|}{a_{\text{max_decel}}}, 1.0\right) & \text{if } a_{\text{veh}}^{y} < 0 \\ 0 & \text{otherwise} \end{cases} \tag{20}$$

Based on the Tesla Model 3 owner's manual (Tesla 2024), $L$ denotes the wheelbase of the vehicle (2.875 m for the simulated Model 3), and $v_{veh}^{x}$ is the current lateral speed ( $\mathrm{m/s}$ ). A small constant $\epsilon = 10^{-6}$ is used to avoid division by zero. The maximum steering angle $\delta_{\max}$ is set to 0.52 radians in the simulation. The maximum acceleration ( $a_{\text{max_accel}}$ ) and deceleration ( $a_{\text{max_decel}}$ ) are approximately 9.25 $\mathrm{m/s^2}$ and 8.88 $\mathrm{m/s^2}$.

For the pedestrian agent, the action $\left(a_{ped}^{x}, a_{ped}^{y}\right)$ is interpreted as changes in acceleration. The pedestrian's velocity is updated using a semi-implicit integration scheme:

$$v_{ped}^{x}(t+1) = v_{ped}^{x}(t) + 0.5 \cdot a_{ped}^{x} \cdot \Delta t \tag{21}$$

$$v_{ped}^{y}(t+1) = v_{ped}^{y}(t) + 0.5 \cdot a_{ped}^{y} \cdot \Delta t \tag{22}$$

where $\Delta t$ is the fixed simulation time step in CARLA. The updated pedestrian position is then computed as:

$$x_{ped}(t+1) = x_{ped}(t) + \left(v_{ped}^{x} + 0.5 \cdot a_{ped}^{x} \cdot \Delta t\right) \cdot \Delta t \tag{23}$$

$$y_{ped}(t+1) = y_{ped}(t) + \left(v_{ped}^{y} + 0.5 \cdot a_{ped}^{y} \cdot \Delta t\right) \cdot \Delta t \tag{24}$$

These updates ensure a realistic and dynamically responsive pedestrian movement, aligning with the reinforcement learning policy's decision-making.

All state–action–reward sequences generated during safety-critical interactions were stored in the simulation replay buffer described above. This buffer enables continuous online reinforcement learning. The MA-SST-DDPG policy adapts over time to different intersection layouts, varied pedestrian paths, and changing vehicle dynamics. During training, mini-batches from the buffer update the actor and critic networks using temporal-difference learning. This process keeps the policy scalable across diverse intersection scenarios.

To guide this learning process effectively, the framework implements a multi-component reward function. Following previous studies (Anzalone *et al.* 2021, Pérez-Gil *et al.* 2022), reinforcement learning in CARLA uses a reward function to guide control. This study designed an online reward function that combines safety constraints with realistic behavior. The reward has three parts: a collision penalty, a goal-reaching reward, and a speed-keeping incentive. This design connects short-term actions to long-term driving objectives:

**Collision Penalty ($R_{\text{collision}}$):** Safety remains the primary constraint during online adaptation. A severe negative reward is assigned immediately when a collision occurs:

$$R_{\text{collision}} = \begin{cases} C_{\text{collision}}, & \text{if a collision occurs} \\ 0, & \text{otherwise} \end{cases} \tag{25}$$

where $C_{\text{collision}}$ is a constant (-200) to strongly discourage unsafe maneuvers (Pérez-Gil *et al.* 2022).

**Goal Achievement Reward ($R_{\text{goal}}$):** To ensure successful task completion, the agent receives a substantial positive reward upon reaching the target location without violations:

$$R_{\text{goal}} = \begin{cases} C_{\text{goal}}, & \text{if the goal is reached} \\ 0, & \text{otherwise} \end{cases} \tag{26}$$

where $C_{\text{goal}}$ is typically set to 100, promoting efficient navigation (Pérez-Gil *et al.* 2022).

To prevent any single objective from dominating the learning signal, the three reward components are integrated through a weighted aggregation scheme. The total reward at each time step is defined as:

$$R(t) = w_1 \cdot R_{\text{collision}} + w_2 \cdot R_{\text{goal}} + w_3 \cdot R_v \tag{27}$$

where $w_1, w_2$, and $w_3$ are non-negative weights satisfying $w_1 + w_2 + w_3 = 1$, assigned to the collision penalty, goal achievement reward, and speed consistency reward, respectively. The weights are updated adaptively at each episode based on the relative magnitude of each component's contribution to the total reward signal:

$$w_k(n+1) = \frac{w_k(n) \cdot \exp(\lambda \cdot |\bar{R}_k(n)|)}{\sum_{j=1}^{3} w_j(n) \cdot \exp\left(\lambda \cdot \left|\bar{R}_j(n)\right|\right)} \tag{28}$$

where $\bar{R}_k(n)$ denotes the mean value of reward component $k$ over episode $n$, and $\lambda > 0$ is a temperature parameter controlling the sensitivity of weight adaptation. This formulation dynamically amplifies the influence of the most active reward component during training. The three components differ fundamentally in activation density, as summarized in Table 4. $R_v$ generates a non-zero signal at every time step, while $R_{\text{collision}}$ and $R_{\text{goal}}$ are event-triggered, evaluating to zero at most time steps and producing non-zero values only upon collision or goal arrival. During normal driving phases, $R_v$ therefore constitutes the sole active signal, anchoring agent kinematics to realistic speed distributions. As CurvTTC decreases, Bellman bootstrapping propagates the anticipated $R_{\text{collision}}$ backward through the critic network, progressively overriding the speed regularization gradient, even though the collision reward itself remains zero until an actual collision occurs.

**Table 4 Per-step activation density and dominant reward signals across interaction phases**

| Interaction Phase | Per-Step Check Result | $r_{\text{speed}}$ (nonzero) | $r_{\text{collision}}$ (non-zero) | $r_{\text{goal}}$ (nonzero) | Effective Dominant Signal |
|---|---|---|---|---|---|
| Conflict approaching (2s< CurvTTC<5 s) | No event triggered | Active | — | — | $r_{\text{speed}}$ |
| Emergency evasion (CurvTTC < 2s) | No event triggered | Active | — | — | $r_{\text{speed}}$ |
| Collision occurs | $r_{\text{collision}}$ triggered | Active | Active | — | $r_{\text{collision}}$ dominant |
| Goal reached | $r_{\text{goal}}$ triggered | Active | — | Active | $r_{\text{goal}}$ dominant |

This temporally structured design preserves behavioral diversity by allowing $R_v$ to function as a behavioral prior during non-critical phases, while $R_{\text{collision}}$ and $R_{\text{goal}}$ govern critical decision boundaries through value bootstrapping. High-intensity evasive responses such as emergency braking or rapid lateral displacement emerge naturally when safety demands override speed regularization, helping the VPSCI dataset capture a broad range of realistic avoidance behaviors. To assess the robustness of the weighting scheme, a sensitivity analysis was conducted under seven fixed weight configurations alongside the adaptive scheme, with collision rate, goal achievement rate, ADE, and FDE recorded at convergence. Results are reported in Table 5.

Table 5 Sensitivity analysis of reward weight configurations

| Configuration | $w_1$ (Collision) | $w_2$ (Goal) | $w_3$ (Speed) | Collision Rate (%) | Goal Achievement Rate (%) | ADE (m) | FDE (m) |
|---|---|---|---|---|---|---|---|
| Equal weights | 0.33 | 0.33 | 0.33 | 18.4 | 71.2 | 0.214 | 0.423 |
| Safety dominant | 0.60 | 0.20 | 0.20 | 12.1 | 74.3 | 0.187 | 0.371 |
| Goal dominant | 0.20 | 0.60 | 0.20 | 21.7 | 78.6 | 0.231 | 0.458 |
| Speed dominant | 0.20 | 0.20 | 0.60 | 15.3 | 72.8 | 0.196 | 0.389 |
| Safety + Goal | 0.45 | 0.45 | 0.10 | 13.8 | 76.4 | 0.179 | 0.354 |
| Safety + Speed | 0.45 | 0.10 | 0.45 | 11.2 | 73.1 | 0.163 | 0.321 |
| Goal + Speed | 0.10 | 0.45 | 0.45 | 23.5 | 77.9 | 0.248 | 0.491 |
| Adaptive (proposed) | - | - | - | 8.3 | 81.2 | 0.072 | 0.142 |

The adaptive scheme performs favorably relative to the fixed configurations across the reported metrics. Safety-dominant and safety-plus-speed settings yield the lowest collision rates among fixed configurations (12.1% and 11.2%), while goal-dominant and goal-plus-speed settings achieve higher goal achievement rates at the cost of substantially elevated collision rates (21.7% and 23.5%). The adaptive scheme achieves the lowest collision rate (8.3%), highest goal achievement rate (81.2%), and lowest trajectory errors (ADE = 0.072 m, FDE = 0.142 m), indicating that the adaptive weighting better accommodates the shifting dominance of reward components across heterogeneous interaction phases in terms of safety and trajectory accuracy.

### 3.3. CARLA-Based Safety-Critical Data Generation

After online learning, the Refined MA-SST-DDPG model was deployed in CARLA to generate a large-scale dataset of vehicle-pedestrian interactions under high-risk conditions. The study defined eight vehicle-pedestrian crossing types to represent diverse interaction scenarios. Each type reflects a pedestrian crossing direction during a vehicle right-turn, as shown in Figure 4 (b). Four scenarios are derived from Location A (① – ④) and four from Location B (⑤ – ⑧). Details of these scenarios are listed in Table 6. Each scenario was run until a minimum of 20,000 qualifying interaction episodes were collected after post-processing. Each episode ran at a fixed time step of 0.05 seconds and was terminated upon collision, goal arrival, or after a maximum of 10 seconds. All simulations were executed on an Intel Core i9-9960X CPU @ 3.10 GHz with dual NVIDIA Quadro RTX 5000 GPUs in synchronous mode, requiring approximately 812 hours of total computation, primarily due to the high rejection rate of raw episodes that did not meet the safety-critical qualification criteria.

**Table 6 Scenario descriptions and dataset files**

| No. | Location | Scenario Description | File Name |
|---|---|---|---|
| ① | A | Vehicle right-turn, pedestrian crossing from South to North | LocationA_PedCross_S-N.csv |
| ② | | Vehicle right-turn, pedestrian crossing from North to South | LocationA_PedCross_N-S.csv |
| ③ | | Vehicle right-turn, pedestrian crossing from East to West | LocationA_PedCross_E-W.csv |
| ④ | | Vehicle right-turn, pedestrian crossing from West to East | LocationA_PedCross_W-E.csv |
| ⑤ | B | Vehicle right-turn, pedestrian crossing from South to North | LocationB_PedCross_S-N.csv |
| ⑥ | | Vehicle right-turn, pedestrian crossing from North to South | LocationB_PedCross_N-S.csv |
| ⑦ | | Vehicle right-turn, pedestrian crossing from East to West | LocationB_PedCross_E-W.csv |
| ⑧ | | Vehicle right-turn, pedestrian crossing from West to East | LocationB_PedCross_W-E.csv |

Figure 5 shows sequential snapshots of the eight scenarios to illustrate how each interaction develops over time. Panels (a) – (d) show interactions at Location A (① – ④), and panels (e) – (h) show interactions at Location B (⑤ – ⑧). Each sequence includes key stages of the right-turn and pedestrian crossing: approach, entry into the conflict zone, the critical interaction, and final clearance.

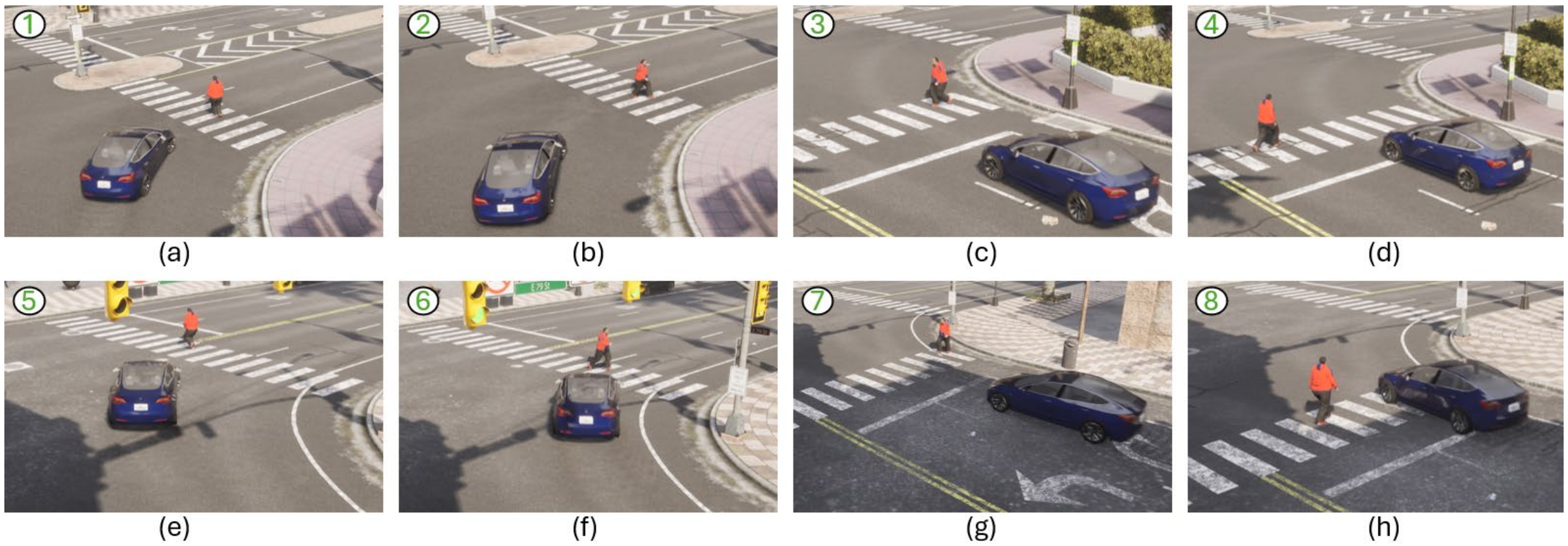

**Figure 5: Sequential snapshots of vehicle–pedestrian interactions in the eight scenarios (① – ⑧) from the CARLA simulation**

All interaction data are recorded at high temporal resolution for each simulation step. By randomizing vehicle initial speeds, pedestrian walking speeds, and spatial configurations, the three-stage framework generated data systematically cover a wide spectrum of interaction patterns. This approach promotes greater diversity in high-risk cases across a range of kinematic conditions and pedestrian responses. Table 7 summarizes the dataset structure and variables recorded for each frame.

**Table 7 Detailed variable definitions of the CARLA-based interaction dataset**

| Category | Variable Name | Description |
|---|---|---|
| Simulation Metadata | count | Interaction episode index |
| | frame | Frame number within the episode |
| Vehicle State | veh_id | Unique vehicle identifier |
| | $d^y_{veh}(t)$, $d^x_{veh}(t)$ | Vehicle position coordinates (meters) |
| | $v^y_{veh}(t)$, $v^x_{veh}(t)$ | Vehicle velocity components in x and y directions (m/s) |
| Pedestrian State | ped_id | Unique pedestrian identifier |
| | $d^y_{ped}(t)$, $d^x_{ped}(t)$ | Pedestrian position coordinates (meters) |
| | $v^y_{ped}(t)$, $v^x_{ped}(t)$ | Pedestrian velocity components in x and y directions (m/s) |
| Relative Motion | distance | Euclidean distance between vehicle and pedestrian (meters) |
| | CurvTTC | Curvilinear Time-to-Collision (seconds) |

A set of post-processing filters is applied to ensure data quality and usability. Each interaction episode (grouped by count ID) is retained if it meets all of these conditions:

(1) Contains minimum CurvTTC <5 s values (episodes with safety-critical cases are included).

(2) All four velocity components ($v^y_{veh}(t)$, $v^x_{veh}(t)$, $v^y_{ped}(t)$, $v^x_{ped}(t)$) have fewer than 10 consecutive missing values.

(3) Episodes contain at least 100 frames.

(4) The vehicle moves more than 10 m and the pedestrian moves more than 4 m between the first and last frames.

The minimum CurvTTC, defined as the smallest CurvTTC value observed throughout an interaction, was extracted to represent the most safety-critical moment of the encounter.

## 4. Results and Discussion

### 4.1. Comparison of Hybrid Learning and Data-Driven Learning Approaches

The training process followed a two-stage design. In Stage 1, the MA-SST-DDPG framework was pre-trained with real-world safety-critical interaction data stored in a replay buffer, consisting of 336 conflicts over 9,443 time steps. This stage enabled agents to acquire human-like collision avoidance strategies. In Stage 2, the pre-trained agents were refined through online reinforcement learning in the CARLA simulator, where they updated their policies under diverse and high-risk scenarios. Based on this setup, two training settings were defined: Data-Driven Learning, which uses only real-world data in a single-stage process, and Hybrid Learning, which combines real-world pre-training with simulation-based online refinement. This distinction allowed evaluation of both model architectures and the added benefit of simulation-based learning compared with real data alone. Table 8 lists the key hyperparameters used in training.

**Table 8 Hyperparameters for training the Refined MA-SST-DDPG**

| Category | Hyperparameter | Value | Description |
|---|---|---|---|
| Buffer and Batch | Buffer capacity | 10000 | Size of replay memory for storing experiences |
| | Batch size | 256 | Samples per training iteration |
| Learning Rates | Learning rate (Actor) | 0.0001 | Step size for updating the policy network |
| | Learning rate (Critic) | 0.0002 | Step size for updating the value network |
| RL Parameters | Discount factor | 0.99 | Rate at which future rewards are discounted |
| | Target update rate | 0.01 | Soft update rate for target networks |
| | Total episodes | 3000 | Total number of training episodes |
| | Reward scaling | 100 | Factor used to scale reward differences in velocity |
| SST Architecture | Hidden dimension | 256 | Dimension of hidden state in SST module |
| | Number of SST layers | 2 | Depth of state-space transformer stack |
| | Number of attention heads | 8 | Multi-head attention configuration |
| | Key dimension | 32 | Dimension for attention key/query projections |
| | SST input sequence length | 10 | Historical time steps |
| Actor Network | Input dimension | 6 | State features per agent |
| | Hidden layers | [64 64] | Two fully connected layers with 64 neurons each |
| | Output dimension | 2 | Acceleration actions (longitudinal lateral) |
| | Activation function | ReLU | Nonlinearity used in hidden layers |
| | Output activation | Linear | Direct action output |
| Critic Network | State processing layer | 16 | First layer for state encoding |
| | Action processing layer | 16 | First layer for joint action encoding |
| | Hidden layers after concat | [64 64] | Two layers with 64 neurons each |
| | Output dimension | 1 | Q-value scalar output |
| | Activation function | ReLU | Nonlinearity in hidden layers |

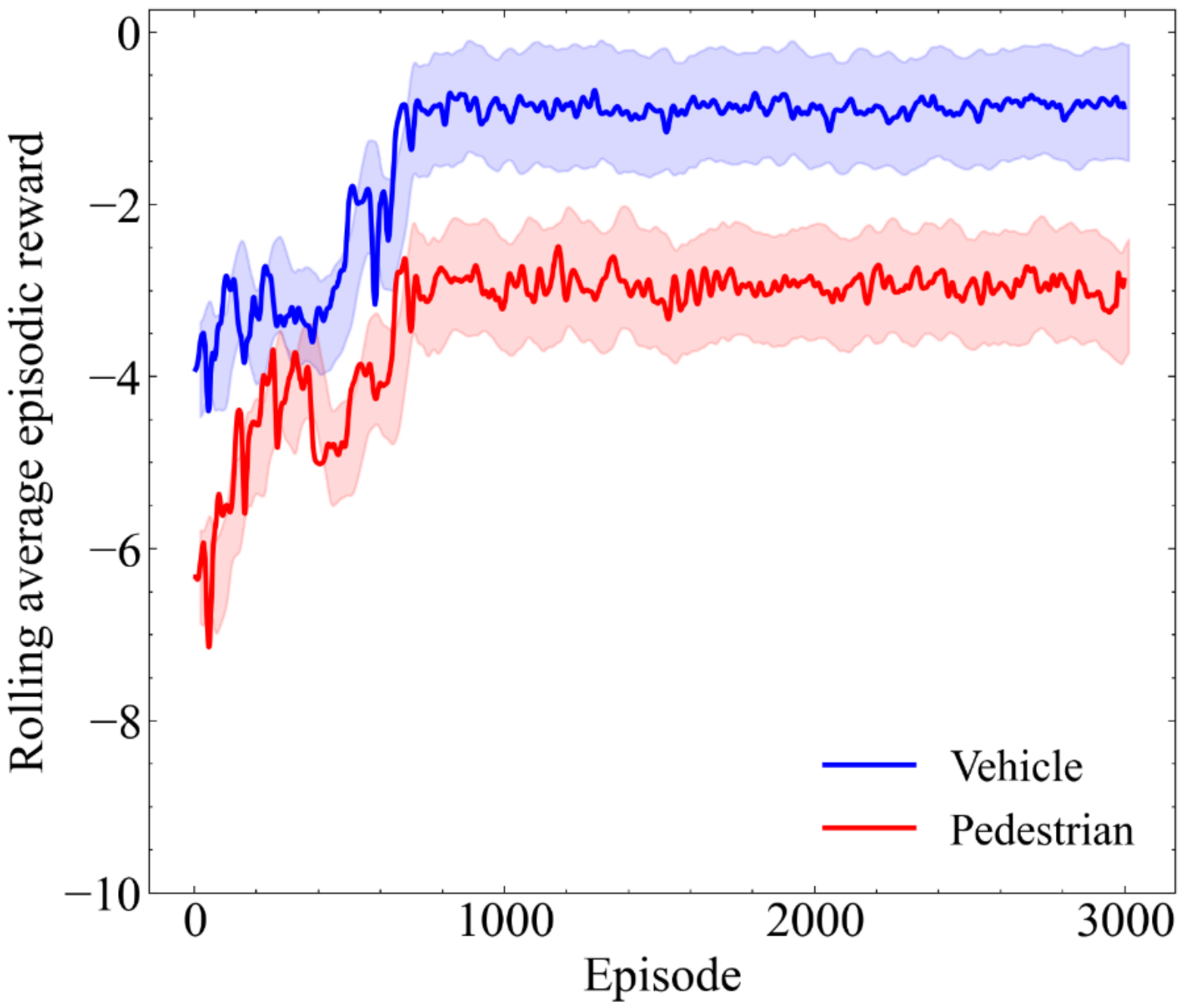

**Figure 6: Rolling average episodic reward of Refined MA-SST-DDPG models during training**

Figure 6 shows the rolling average episodic rewards for both agents during online learning. The results highlight clear differences in how the two agents learn. The vehicle agent exhibits rapid improvement in the early phase of training. Its rewards increase rapidly during the early training stage, indicating that the vehicle agent quickly learns effective collision avoidance strategies. After approximately 800 episodes, the rewards stabilize, suggesting that the vehicle agent has converged to a stable and effective policy. Both agents reach stable policies by episode 1500 with negligible reward fluctuation thereafter. The concurrent stabilization of both reward trajectories, rather than one agent improving at the expense of the other, provides empirical confirmation of the cooperative Nash equilibrium structure, ruling out zero-sum competitive dynamics (Lowe et al. 2017).

The pedestrian agent shows more variable learning behavior. Although it starts with a similar reward level to the vehicle agent, its rewards fluctuate more and eventually stabilize around -3.5. This variability and lower final performance result from the unpredictable nature of pedestrian behavior. Sudden changes in direction, irregular crossing patterns, and adaptive responses to vehicles make the learning process more complex. The performance gap between the two agents reflects differences in their behavior and decision-making. Vehicle movement is structured and follows traffic rules and physical limits. Pedestrian actions are more random, which makes policy learning more challenging. To confirm convergence, the training process monitors the policy gradient and considers learning complete when the gradient norm satisfies $\|\nabla_\theta J(\theta)\|_2 < 10^{-4}$ for 100 consecutive episodes. This result shows that the three-stage framework allows both agents to develop stable and realistic behaviors, which can be used to generate reliable safety-critical interaction data.

This study used the Root Mean Square Error (RMSE) metric to measure the discrepancy between predicted and actual values across various indicators. RMSE is computed as:

$$RMSE = \sqrt{\frac{1}{n}\sum_{i=1}^{n} (\hat{x}_i - x_i)^2} \tag{29}$$

Where $\hat{x}_i$ and $x_i$ are the predicted and observed values, respectively.

In addition, two trajectory-based metrics were adopted: Average Displacement Error (ADE) and Final Displacement Error (FDE). ADE measures the mean Euclidean distance between predicted and observed trajectories over all time steps, while FDE measures the Euclidean distance at the final predicted step. These metrics are defined as:

$$ADE = \frac{1}{nT}\sum_{i=1}^{n}\sum_{t=1}^{T} \left\|\hat{y}_{i,t} - y_{i,t}\right\|_2 \tag{30}$$

$$FDE = \frac{1}{n}\sum_{i=1}^{n} \left\|\hat{y}_{i,T} - y_{i,T}\right\|_2 \tag{31}$$

where $n$ is the number of trajectories, $T$ is the prediction horizon, and $\hat{y}_{i,t}$ and $y_{i,t}$ are the predicted and ground-truth positions for trajectory $i$ at time step $t$.

Direct comparison with existing safety-critical scenario generation methods is constrained by a fundamental difference in problem scope. The predominant body of state-of-the-art work in this domain, including TrafficSim (Suo *et al.* 2021), BITS (Xu *et al.* 2023), Li *et al.* (2023), CCDiff (Lin *et al.* 2025), ChatScene (Zhang *et al.* 2024), and DiffScene (Xu *et al.* 2025), focuses exclusively on vehicle-vehicle interactions such as lane-changing, merging, and cut-in maneuvers. These frameworks optimize for multi-agent trajectory realism within structured lane configurations and do not model pedestrian evasive behavior as a learning objective. Among the limited studies addressing vehicle-pedestrian interactions, existing approaches either employ scripted pedestrian trajectories without adaptive decision-making, or use cognitively inspired behavioral models limited to single-crossing adversarial testing without large-scale dataset generation (Wang *et al.* 2026a). Recent surveys on scenario generation have explicitly identified insufficient coverage of pedestrian and vulnerable road user scenarios as a persistent gap in the field (Ding *et al.* 2023). These structural differences render direct quantitative comparison methodologically inappropriate, and the baselines in Table 9 therefore represent the closest comparable alternatives.

Table 9 reports quantitative results for the Refined MA-SST-DDPG and the comparison models evaluated using RMSE, ADE, and FDE. The 336 HDI safety-critical interactions were partitioned into an 80/20 split, with 269 interactions for training and 67 for held-out evaluation. Two learning paradigms are compared: Data-Driven Learning, trained exclusively on HDI data, and Hybrid Learning (denoted as Refined), combines Stage 1 HDI pre-training with Stage 2 online refinement in CARLA. Both paradigms were evaluated on the same 67 held-out interactions, with all models implemented under identical software, hardware, and hyperparameter search conditions to ensure fair comparison.

The Refined MA-SST-DDPG achieved the best performance on almost all indicators, with velocity errors as low as 0.054 m/s ($\mathbf{v}_{\mathbf{veh}}^{\mathbf{y}}$) and 0.053 m/s ($\mathbf{v}_{\mathbf{ped}}^{\mathbf{y}}$), and the lowest trajectory-level metrics (ADE = 0.072 m, FDE = 0.142 m). In comparison, other multi-agent models such as Refined MA-Transformer-DDPG and Refined MA-DDPG yielded larger errors (ADE = 0.094–0.157 m, FDE = 0.194–0.305 m), confirming the advantage of the SST module in temporal feature selection and its ability to respond to critical motion states. The superiority of the Refined MA-SST-DDPG becomes more evident when compared with single-agent reinforcement learning methods: the best single-agent model, Refined Transformer-DDPG, produced ADE = 0.257 m, FDE = 0.505 m, nearly four times larger than those of the Refined MA-SST-DDPG, while Refined DDPG performed even worse with ADE = 0.623 m, FDE = 1.273 m. Trajectory prediction models such as Refined DenseTNT, MultiPath++, and Social-LSTM also exhibited substantially higher errors, with ADE values between 0.404–0.651 m and FDE values between 0.805–1.330 m, more than five times larger than those of the Refined MA-SST-DDPG. Supervised learning models produced the weakest results; the Refined Transformer, LSTM, and NN models all reported ADE values above 1.2 m and FDE values above 2.2 m. These comparisons indicate that the Refined MA-SST-DDPG is the most accurate among the evaluated model families.

The results also highlight the benefits of the Hybrid Learning approach over Data-Driven Learning. For multi-agent models, the Refined MA-SST-DDPG reduced ADE/FDE from 0.078 m / 0.156 m to 0.072 m /

0.142 m, while MA-Transformer-DDPG and MA-DDPG exhibited similar gains (e.g., ADE decreased from 0.098 m to 0.094 m, and from 0.164 m to 0.157 m, respectively). Single-agent models showed consistent improvements as well, with Transformer-DDPG reducing ADE/FDE from 0.271 m / 0.542 m under Data-Driven Learning to 0.257 m / 0.505 m with Hybrid Learning. Even the trajectory prediction and supervised learning models benefited from the Hybrid Learning. For example, DenseTNT reduced its FDE from 0.852 m to 0.805 m, while the supervised Transformer decreased from 2.398 m to 2.205 m.

In summary, the Hybrid Learning consistently yielded lower errors, highlighting the importance of combining real-world pre-training with simulation-based online learning. The Refined MA-SST-DDPG, developed using this hybrid design, achieved stable learning, reduced prediction errors, and stronger generalization to complex vehicle–pedestrian interactions. When compared across the multi-agent reinforcement learning, single-agent reinforcement learning, trajectory prediction, and supervised learning model families, the real-world pre-training and simulation-based online learning framework offers a scalable strategy to improve model performance in other domains where real data is scarce but high-risk scenarios are critical.

**Table 9 Quantitative evaluation of models.**

| Model Type | Training Setting | Model | RMSE $v_{veh}^{y}$ (m/s) | RMSE $v_{veh}^{x}$ (m/s) | RMSE $v_{ped}^{y}$ (m/s) | RMSE $v_{ped}^{x}$ (m/s) | RMSE $d_{vp}^{y}$ (m) | RMSE $d_{vp}^{x}$ (m) | RMSE $d_{pv}^{y}$ (m) | RMSE $d_{pv}^{x}$ (m) | ADE (m) | FDE (m) |
|---|---|---|---|---|---|---|---|---|---|---|---|---|
| Multi-Agent Reinforcement Learning | Hybrid Learning | Refined MA-SST-DDPG | 0.054* | 0.024 | 0.053* | 0.031 | 0.096* | 0.114* | 0.102* | 0.112* | 0.072* | 0.142* |
| | | Refined MA-Transformer-DDPG | 0.056 | 0.024 | 0.056 | 0.035 | 0.134 | 0.134 | 0.136 | 0.133 | 0.094 | 0.194 |
| | | Refined MA-DDPG | 0.133 | 0.041 | 0.131 | 0.043 | 0.312 | 0.172 | 0.218 | 0.188 | 0.157 | 0.305 |
| | Data-Driven Learning | MA-SST-DDPG | 0.058 | 0.023* | 0.056 | 0.030* | 0.098 | 0.118 | 0.103 | 0.126 | 0.078 | 0.156 |
| | | MA-Transformer-DDPG | 0.061 | 0.026 | 0.057 | 0.035 | 0.145 | 0.136 | 0.146 | 0.135 | 0.098 | 0.196 |
| | | MA-DDPG | 0.147 | 0.045 | 0.134 | 0.048 | 0.322 | 0.184 | 0.227 | 0.203 | 0.164 | 0.328 |
| Single-Agent Reinforcement Learning | Hybrid Learning | Refined Transformer- | 0.249 | 0.075 | 0.229 | 0.085 | 0.363 | 0.386 | 0.431 | 0.324 | 0.257 | 0.505 |
| | | Refined DDPG | 0.557 | 0.306 | 0.731 | 0.182 | 0.914 | 0.961 | 0.599 | 0.622 | 0.623 | 1.273 |
| | Data-Driven Learning | Transformer-DDPG | 0.264 | 0.079 | 0.233 | 0.086 | 0.383 | 0.387 | 0.434 | 0.346 | 0.271 | 0.542 |
| | | DDPG | 0.594 | 0.313 | 0.802 | 0.181 | 1.001 | 1.050 | 0.656 | 0.671 | 0.679 | 1.357 |
| Trajectory Prediction | Hybrid Learning | Refined DenseTNT | 0.293 | 0.185 | 0.339 | 0.199 | 0.430 | 0.365 | 0.471 | 0.409 | 0.404 | 0.805 |
| | | Refined MultiPath++ | 0.315 | 0.178 | 0.365 | 0.216 | 0.480 | 0.401 | 0.525 | 0.422 | 0.424 | 0.871 |
| | | Refined Social- | 0.421 | 0.247 | 0.528 | 0.317 | 0.690 | 0.508 | 0.673 | 0.566 | 0.651 | 1.330 |
| | Data-Driven Learning | DenseTNT | 0.315 | 0.185 | 0.353 | 0.212 | 0.454 | 0.383 | 0.488 | 0.414 | 0.422 | 0.852 |
| | | MultiPath++ | 0.343 | 0.195 | 0.383 | 0.237 | 0.497 | 0.415 | 0.525 | 0.441 | 0.453 | 0.923 |
| | | Social-LSTM | 0.452 | 0.250 | 0.524 | 0.318 | 0.688 | 0.556 | 0.712 | 0.588 | 0.654 | 1.353 |
| Supervised Learning | Hybrid Learning | Refined Transformer | 1.075 | 0.422 | 1.592 | 0.257 | 1.742 | 1.092 | 1.554 | 1.552 | 1.203 | 2.205 |
| | | Refined LSTM | 2.051 | 1.848 | 1.951 | 1.864 | 1.942 | 1.832 | 2.033 | 1.998 | 1.771 | 2.551 |
| | | Refined NN | 2.176 | 2.261 | 2.138 | 2.216 | 2.045 | 2.139 | 1.911 | 1.951 | 2.586 | 3.989 |
| | Data-Driven Learning | Transformer | 1.156 | 0.438 | 1.661 | 0.261 | 1.751 | 1.169 | 1.549 | 1.624 | 1.199 | 2.398 |
| | | LSTM | 2.066 | 1.877 | 2.059 | 2.011 | 2.024 | 1.847 | 2.128 | 2.012 | 1.783 | 2.566 |
| | | NN | 2.252 | 2.256 | 2.145 | 2.382 | 2.195 | 2.145 | 2.055 | 2.008 | 2.677 | 3.955 |

* The best values among all the models. ADE and FDE values are averaged over both vehicle and pedestrian evaluation sequence

## 4.2. Generate Safety-Critical Data

This section summarizes the key statistics of the VPSCI dataset, which was generated using the three-stage framework across eight distinct interaction scenarios (see Table 10, Figure 5, and Figure 7). These scenarios, saved as CSV files, capture right-turn vehicle movements interacting with crossing pedestrians at two intersections (Location A and B), with four crossing directions at each location.

Each scenario contains over 20,000 interaction episodes, totaling 198,157 episodes across the dataset. These interactions cover about 3,603 km of vehicle trajectories and 2,625 km of pedestrian trajectories, reflecting diverse motion patterns and spatial layouts. Table 10 summarizes interaction counts and trajectory lengths for each file.

**Table 10 Trajectory Statistics of CARLA-Based Vehicle-Pedestrian Interaction Dataset**

| No. | File Name | Interaction Count | Vehicle trajectory length (km) | Pedestrian trajectory length (km) |
|---|---|---|---|---|
| ① | LocationA_PedCross_S-N.csv | 20066 | 392.1 | 285.1 |
| ② | LocationA_PedCross_N-S.csv | 30533 | 543.5 | 374.6 |
| ③ | LocationA_PedCross_E-W.csv | 25066 | 471.8 | 335.2 |
| ④ | LocationA_PedCross_W-E.csv | 23055 | 439.2 | 312.4 |
| ⑤ | LocationB_PedCross_S-N.csv | 29729 | 498.1 | 372.2 |
| ⑥ | LocationB_PedCross_N-S.csv | 24702 | 431.9 | 322.5 |
| ⑦ | LocationB_PedCross_E-W.csv | 23504 | 422.9 | 311.8 |
| ⑧ | LocationB_PedCross_W-E.csv | 21502 | 403.6 | 311.2 |

To illustrate the temporal and spatial dynamics, Figure 7 presents selected interaction trajectories from the eight right-turn scenarios in the VPSCI dataset, with scenarios ① - ④ from Location A and ⑤ - ⑧ from Location B. Vehicle paths are shown in blue and pedestrian paths in red, with color gradients encoding instantaneous speed (0–8 m/s); darker shades indicate higher speeds, while lighter shades approaching white denote near-zero speed, signaling deceleration or a complete pause that reflects an active yielding maneuver. Green circles mark agent starting positions, and the directional arrows indicate the corresponding direction of motion. By jointly examining the trajectory shape and the speed-encoded color gradient of both agents, the yielding party in each interaction can be identified: when the vehicle trajectory contains a near-white segment while the pedestrian maintains its speed, the vehicle is yielding; conversely, when the pedestrian trajectory fades to white while the vehicle continues through, the pedestrian is yielding. The yielding outcome of each scenario is labeled in the corresponding subplot (Vehicle-yield or Pedestrian-yield). These trajectories demonstrate the geometric diversity and behavioral fidelity of the dataset across varied crossing configurations, supporting its applicability for training and evaluating interactive collision avoidance models.

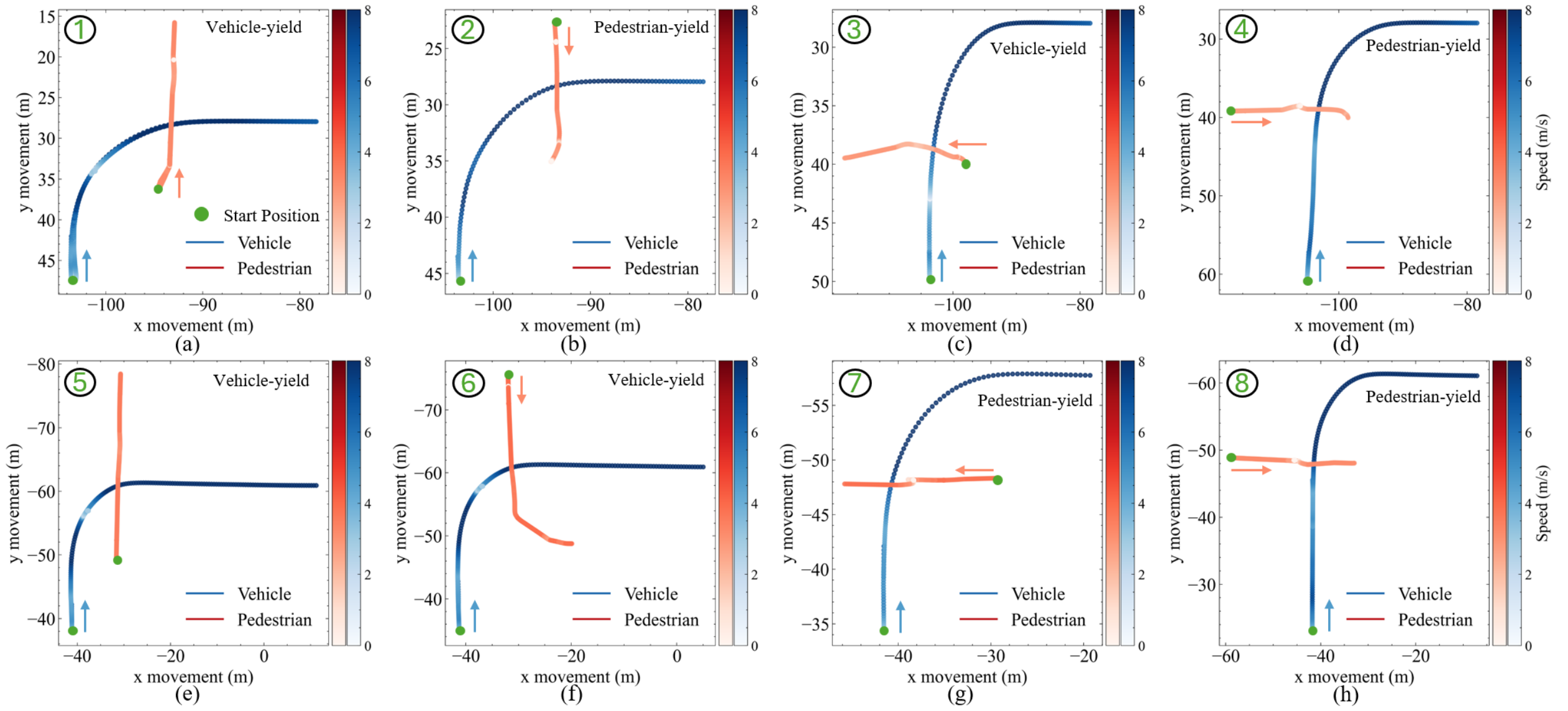

**Figure 7: Velocity-mapped trajectories for scenarios ① – ⑧**

To validate the VPSCI dataset's representativeness, two real-world datasets were selected for comparison. Argoverse 2 (Wilson *et al.* 2023) contains large-scale urban driving data from six U.S. cities with diverse vehicle-pedestrian interactions. TGSIM (Talebpour *et al.* 2024) captures trajectory data from highway and urban environments in Chicago and Washington, D.C. using aerial and infrastructure cameras. For each interaction, the minimum CurvTTC was extracted to represent the most safety-critical moment of the encounter. The results show that 100% of VPSCI interactions fall below 5 s, indicating a complete focus on high-risk vehicle–pedestrian encounters. In contrast, only 25.18% of Argoverse 2 and 55.77% of TGSIM interactions exhibit CurvTTC below 5 s, with most samples corresponding to routine, low-risk traffic events. This distinction highlights VPSCI's role as a targeted testbed for evaluating autonomous driving safety performance under extreme conditions.

The VPSCI dataset is available for download at https://github.com/Qpu523/VPSCI-Dataset. The repository also includes 144 complementary videos, providing visual documentation of the vehicle-pedestrian interactions. These videos enable researchers to conduct qualitative analysis and verify behavioral patterns, enhancing the dataset's utility beyond the quantitative trajectory data alone.

### 4.3. Behavioral Realism Validation

Validating behavioral realism requires distributional analysis beyond geometric trajectory matching, since generated data may approximate average motion paths while failing to capture the kinematic variability of naturalistic decision-making. This section compares the real-world HDI dataset and the synthesized VPSCI dataset along two complementary dimensions: the temporal severity of conflicts (CurvTTC) and the behavioral response to those conflicts (pedestrian yielding probability).

Figure 8 shows the CurvTTC distributions for (a) the HDI dataset and (b) the VPSCI dataset, displayed as half-violin kernel density estimates paired with horizontal density histograms; the red line marks the median and the two black lines mark Q1 and Q3.

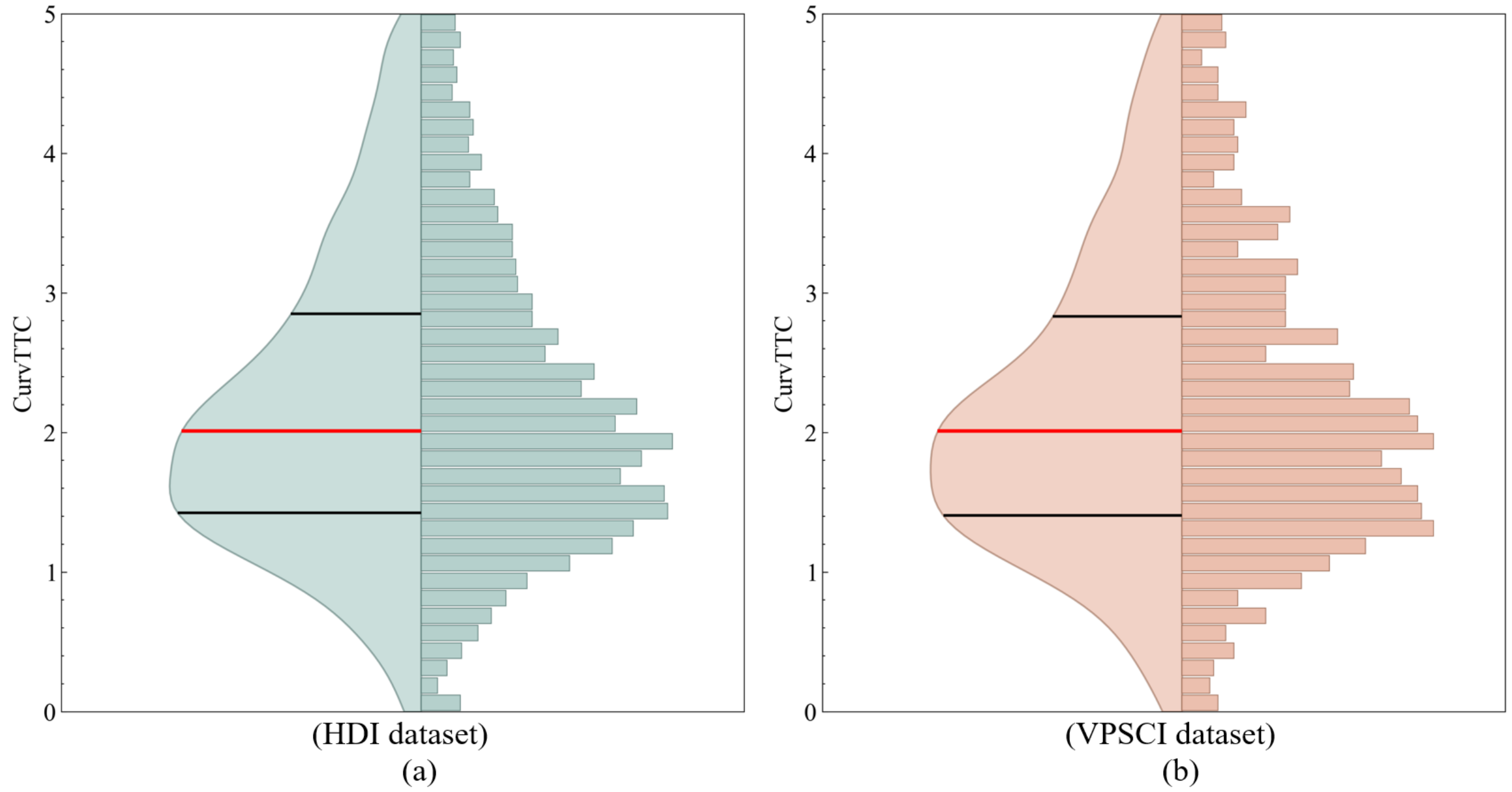

**Figure 8 : Comparison of CurvTTC distributions between (a) real-world HDI dataset and (b) VPSCI dataset**

Both distributions are unimodal with the bulk of conflicts in the 1–3 s range and a peak near 2 s. The quartile statistics are nearly identical: HDI yields Q1 = 1.43 s, median = 2.01 s, Q3 = 2.85 s, while VPSCI yields Q1 = 1.41 s, median = 2.01 s, Q3 = 2.83 s. The Kolmogorov–Smirnov test (Berger and Zhou 2014) returned D = 0.0149 with p = 0.9942, detecting no statistically significant difference. Because a non-significant test does not establish equivalence, a bootstrap equivalence test was conducted on the Wasserstein distance (0.0150 s), whose 95% confidence interval of [0.025, 0.090] s fell below the predefined 0.1 s margin, the temporal resolution of the trajectory data, formally supporting distributional equivalence within ±0.1 s.

A parallel procedure was applied to characterize pedestrian yielding behavior. The first frame where CurvTTC ≤ 5 s was marked as the onset, at which the initial pedestrian speed and vehicle–pedestrian distance were recorded. A pedestrian was classified as yielding if the pedestrian stopped or substantially reduced walking speed to allow the vehicle to pass through the conflict zone first, rather than continuing the crossing. All trajectories were manually checked to ensure correct classification. For each bin in the joint space of vehicle–pedestrian distance and initial pedestrian speed, the pedestrian yielding probability was computed as the percentage of yielding pedestrians among all interactions falling into that bin.

Figure 9 shows the resulting pedestrian yielding probability surfaces over the joint space of vehicle–pedestrian distance and initial pedestrian speed, with iso-probability contours (20%, 40%, 60%, 80%) overlaid; warmer colors indicate higher yielding likelihood.

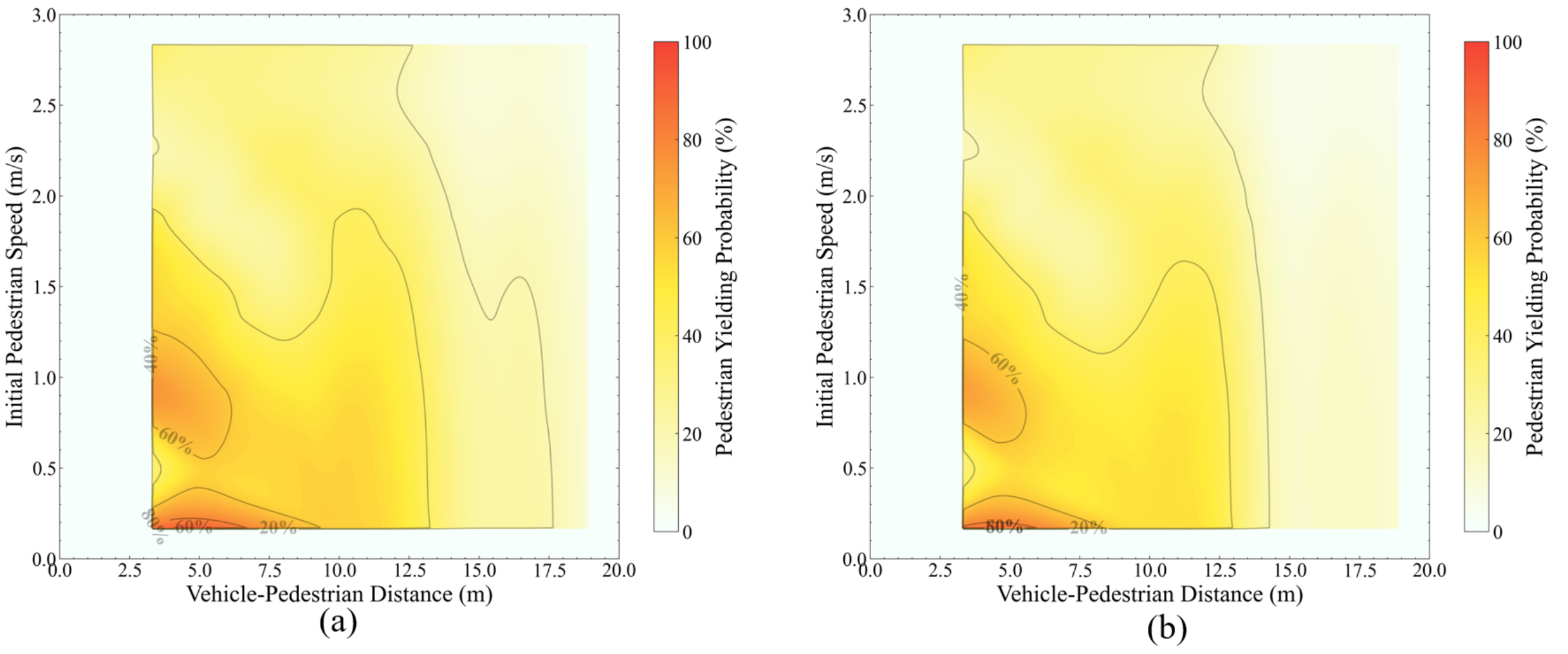

**Figure 9 Comparison of pedestrian yielding probability between (a) real-world HDI dataset and (b) VPSCI dataset**

Both surfaces exhibit close qualitative agreement. A dominant high-probability region (>80%) emerges in the lower-left corner, where pedestrians approach at low speed (≤ 0.5 m/s) within close range (3 – 5 m)—a configuration in which pedestrians, already nearly stationary near an oncoming vehicle, are most likely to yield. Yielding probability decays monotonically as distance increases or pedestrian speed rises, with the 40% and 60% contours occupying nearly identical positions in both subplots. A secondary high-probability region near (distance ≈ 4 – 6 m, speed ≈ 0.8 – 1.2 m/s) is also reproduced, while both surfaces converge to <20% in the far-distance, high-speed regime (distance > 13 m, speed > 1.5 m/s) where pedestrians are committed to completing the crossing.

Taken together, the CurvTTC alignment and the matching pedestrian yielding decision boundaries demonstrate that the VPSCI dataset preserves both the kinematic statistics of real conflicts and the behavioral logic governing pedestrian responses to oncoming vehicles.

### 4.4. Turing Test

A Turing test-inspired perceptual study was conducted to assess the human-likeness of collision avoidance behaviors generated by the three-stage framework. Four safety-critical vehicle-pedestrian scenarios were selected from the CARLA simulated environment. For each scenario, three videos were generated: one from the default CARLA controller, one from the three-stage framework generated scenario, and one using mapped real-world trajectories. This produced 12 videos (4 per type), all rendered from a consistent bird's-eye view in CARLA. Real-world videos were created by projecting naturalistic trajectory data into CARLA with minimal spatial alignment to preserve original motion properties. Videos were generated by the three-stage framework under the same initial conditions. The CARLA baseline videos used default simulator behaviors without any learned policy.

A total of 51 participants (60% male, age range: 20–45 years) evaluated the realism of vehicle–pedestrian interactions across 12 videos. All participants have backgrounds in transportation engineering or related field, and were unaware of the study hypotheses to minimize response bias. Each set included four safety-critical scenarios, with one video from each source: CARLA baseline, three-stage framework, and real-world projection. All videos were shown from a bird's-eye view and anonymized to remove labels or cues. To reduce order bias, the 12 videos were randomized for each participant. After watching each video, participants answered: "How realistically does the vehicle–pedestrian interaction in this video reflect real-world behavior?" Responses used a 5-point Likert scale (1 = Very unrealistic, 5 = Very realistic) to measure perceived naturalness based on motion dynamics. Representative video examples of the evaluated interactions are publicly available at https://www.youtube.com/watch?v=yxsmcCsbB1Y.

This section analyzed the consistency of each evaluator's scores to ensure reliable statistical comparisons. Most evaluators displayed moderate means and reasonable variation, suggesting consistent and discriminative ratings. A few outliers were noted. Subject #16 assigned uniformly high scores with near-zero variance, showing no differentiation. Subject #2 provided very low and highly variable scores, indicating possible bias or misunderstanding. A small group, including subjects #4, #5, #12, #13, and #21, showed high ratings with minimal variance. Seven participants were excluded because their response patterns showed inattentiveness or systematic rating bias instead of genuine evaluation of behavioral realism.

To check consistency among the remaining participants, this research calculated the intraclass correlation coefficient (ICC) using a two-way random effects model for absolute agreement (ICC[2,k]) (Koo and Li 2016). The ICC for the 44 valid participants was 0.763 (95% CI: [0.590, 0.830]), which indicates good reliability. This result shows that participants had substantial agreement in their judgments of behavioral realism, supporting the reliability of the evaluation protocol. After these steps, the final analysis included 44 participants, ensuring that the results reflected attentive and accurate human judgments.

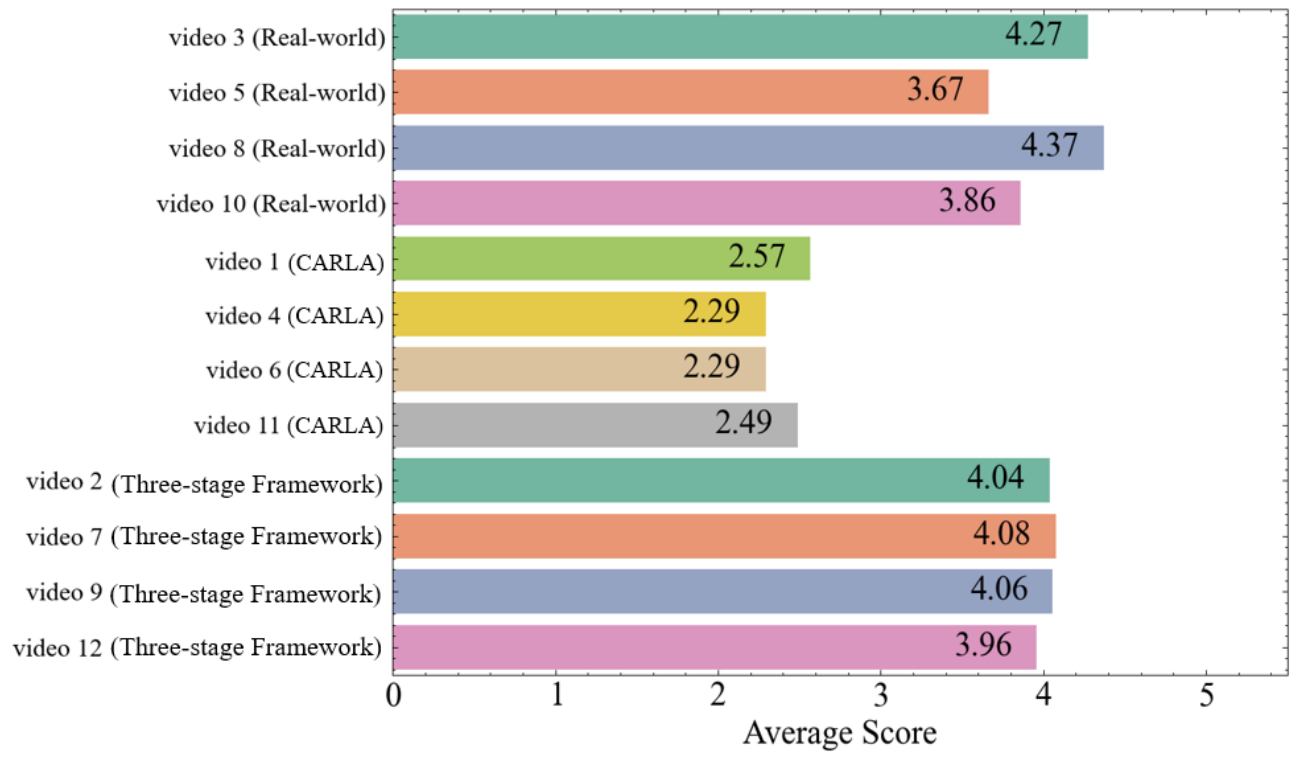

(a)

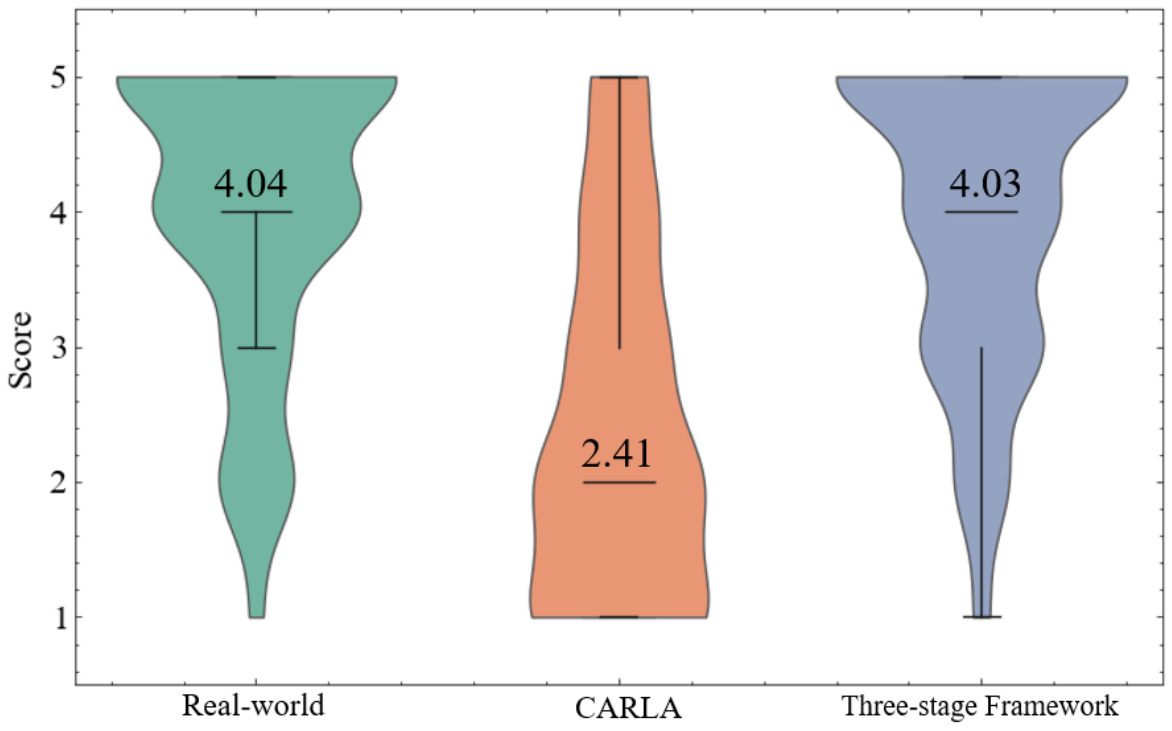

(b)

**Figure 10: (a) Average scores per video; (b) Score distributions by video type in the Turing test**

After removing non-discriminative evaluators, this research aggregated the remaining ratings to compare perceived realism across video types. Figure 10 (a) shows that real-world videos (3, 5, 8, 10) received the highest scores, with video 8 reaching 4.37. Videos generated by the three-stage framework (2, 7, 9, 12) also scored high (3.96–4.08), suggesting participants viewed them as nearly equivalent to real-world interactions.

In contrast, CARLA autopilot videos (1, 4, 6, 11) scored much lower (2.29–2.57), indicating poor behavioral fidelity.

Figure 10 (b) shows the score distribution for each video type. Real-world videos and videos generated by the three-stage framework had similarly high medians near 4 with narrow spreads, indicating consistent perception of natural behavior. CARLA autopilot videos showed lower and more dispersed scores, suggesting poor realism. The sharp clustering of three-stage framework scores near the upper end reflects low variability and supports the model's ability to generate human-like interactions.

Because each participant rated videos from all three sources under the same protocol, the Turing-test data constitutes a repeated-measures (within-subjects) design rather than three independent samples. To account for this dependency, this study applied paired-samples t-tests on subject-level mean scores, calculated effect sizes using Cohen's $d_z$ (Goulet-Pelletier and Cousineau 2018), and employed the Two One-Sided Tests (TOST) (Lauzon and Caffo 2009) procedure, adapted for paired data, to assess equivalence. The paired t-test was used to evaluate whether the mean perceptual scores between any two video types differed significantly within the same participants:

$$t = \frac{\bar{d}}{s_d/\sqrt{n}}, df = n - 1 \tag{32}$$

where $d_i = x_{1i} - x_{2i}$ is the within-subject difference between the two conditions for participant $i$, and $\bar{d}$ and $s_d$ are the mean and standard deviation of the paired differences across the $n$ participants. A p-value < 0.05 indicates a statistically significant difference.

Cohen's $d_z$ was calculated to quantify the magnitude of each within-subject difference. Unlike the pooled-variance Cohen's d for independent samples, $d_z$ standardizes the mean paired difference by the standard deviation of the paired differences, which is the appropriate effect size for repeated-measures comparisons:

$$d_z = \frac{\bar{d}}{s_d} \tag{33}$$

To formally test for equivalence, this study employed the TOST procedure. This method flips the null hypothesis to test if the difference between two groups is small enough to be considered practically equivalent. This study defined an equivalence margin ( $\Delta$ ) of $\pm 0.5$ on the 5-point rating scale. The null hypotheses for TOST are that the true difference is greater than the upper bound $(\mu_1 - \mu_2 \geq \Delta)$ or less than the lower bound $(\mu_1 - \mu_2 \leq -\Delta)$. Two separate one-sided t-tests are conducted on the paired differences:

$$t_1 = \frac{\bar{d} - \Delta}{s_d/\sqrt{n}}, t_2 = \frac{\bar{d} + \Delta}{s_d/\sqrt{n}}, df = n - 1 \tag{34}$$

If both one-sided tests reject their respective null hypotheses (both $p < 0.05$ ), statistical equivalence can be concluded within the margin of $\pm\Delta$.

Table 11 Statistical comparison of perceptual scores between video types

| **Group 1** | **Group 2** | Mean Diff | p-value | Cohen's $d_z$ | 95% CI of Diff | TOST p (Δ= 0.5) | Conclusion |
|---|---|---|---|---|---|---|---|
| CARLA | Three-stage Framework | -1.623 | <0.001 | -1.31 | [-1.97, -1.27] | 1.00 | Different |
| CARLA | Real-world | -1.632 | <0.001 | -1.26 | [-2.00, -1.27] | 1.00 | Different |
| Three-stage Framework | Real-world | -0.010 | 0.909 | -0.016 | [-0.181, 0.162] | <0.001 | Equivalent |

The results, summarized in Table 11, show that the default CARLA controller produced behaviors that were perceived as significantly less natural than both three-stage framework and real-world videos. This perceptual difference is consistent with the behavioral limitation of CARLA's rule-based Traffic Manager, which applies fixed proximity-based braking thresholds that are insufficient to generate adaptive evasive responses when pedestrians enter the conflict zone, resulting in either delayed reactions or unrealistically abrupt stops that human evaluators identified as artificial. This is evidenced by the large, statistically significant within-subject mean differences (both $p < 0.001$) with large paired effect sizes (Cohen's $d_z = -1.31$ and $-1.26$). In contrast, the comparison between the three-stage framework and real-world videos provides strong evidence for perceptual equivalence: the paired difference was negligible ($d_z = -0.016$, $p = 0.909$), its 95% confidence interval ([−0.181, 0.162]) fell entirely within the predefined equivalence margin of ±0.5, and the paired TOST procedure was significant ($p < 0.001$), formally rejecting the hypothesis of a meaningful difference. The linear mixed-effects model on all per-video ratings corroborated these findings, yielding a non-significant contrast between the three-stage framework and real-world videos ($\beta = 0.010$, SE = 0.100, $p = 0.922$) alongside large significant deficits for the CARLA baseline (both $p < 0.001$). This allows the conclusion that the scores for the three-stage framework and real-world data are statistically equivalent within the predefined margin of ±0.5 points. These results validate the three-stage framework's ability to replicate human-like vehicle–pedestrian interactions. Its trajectories were perceptually indistinguishable from real-world data, supporting its use in generating naturalistic safety-critical scenarios for modeling and evaluation.

### 4.5. Conflict Rate Analysis

This section examines the relationship between initial kinematic conditions and conflict occurrence using the generated safety-critical scenarios, and evaluates whether the three-stage framework produces speed-risk relationships that align with real-world pedestrian-vehicle dynamics. To ensure the findings are not an artifact of a single indicator, conflict rates were computed using two complementary surrogate safety measures: CurvTTC, a process-based indicator capturing collision-course severity, and PET, an event-based indicator widely used for vehicle-pedestrian interactions.

Each interaction was analyzed to identify the moment of risk emergence. The first frame where CurvTTC $\leq 5$ s was marked as the interaction onset, at which the vehicle and pedestrian speeds were recorded as the initial speeds. Under the CurvTTC criterion, an interaction was labeled as a conflict if CurvTTC ≤ 2 s at any frame (Jian *et al.* 2018, Pu *et al.* 2026b). Under the PET criterion, PET was computed as the time gap between the moment the first road user left the conflict area and the moment the second entered it. Interactions with PET ≤ 5 s were included in the analysis (Fu *et al.* 2016, Islam *et al.* 2023), and an interaction was labeled as a conflict if PET ≤ 1.2 s, consistent with the dangerous-conflict threshold for vehicle-pedestrian interactions proposed by Fu *et al.* (2016). Interactions in which the two road users did not successively traverse a common conflict area, typically because timely evasive maneuvers prevented the trajectories from crossing, yield undefined PET values; as a result, the PET-based analysis identified 21.3% fewer conflicts than the CurvTTC criterion, reflecting the different measurement scopes of the two indicators. Initial vehicle and pedestrian speeds were discretized into bins from 0 to 4 m/s in 0.5 m/s intervals, and the conflict rate in each bin was computed as the percentage of conflict-labeled interactions. For comparison, interactions generated by CARLA's default autopilot under the same settings were processed with the identical procedure.

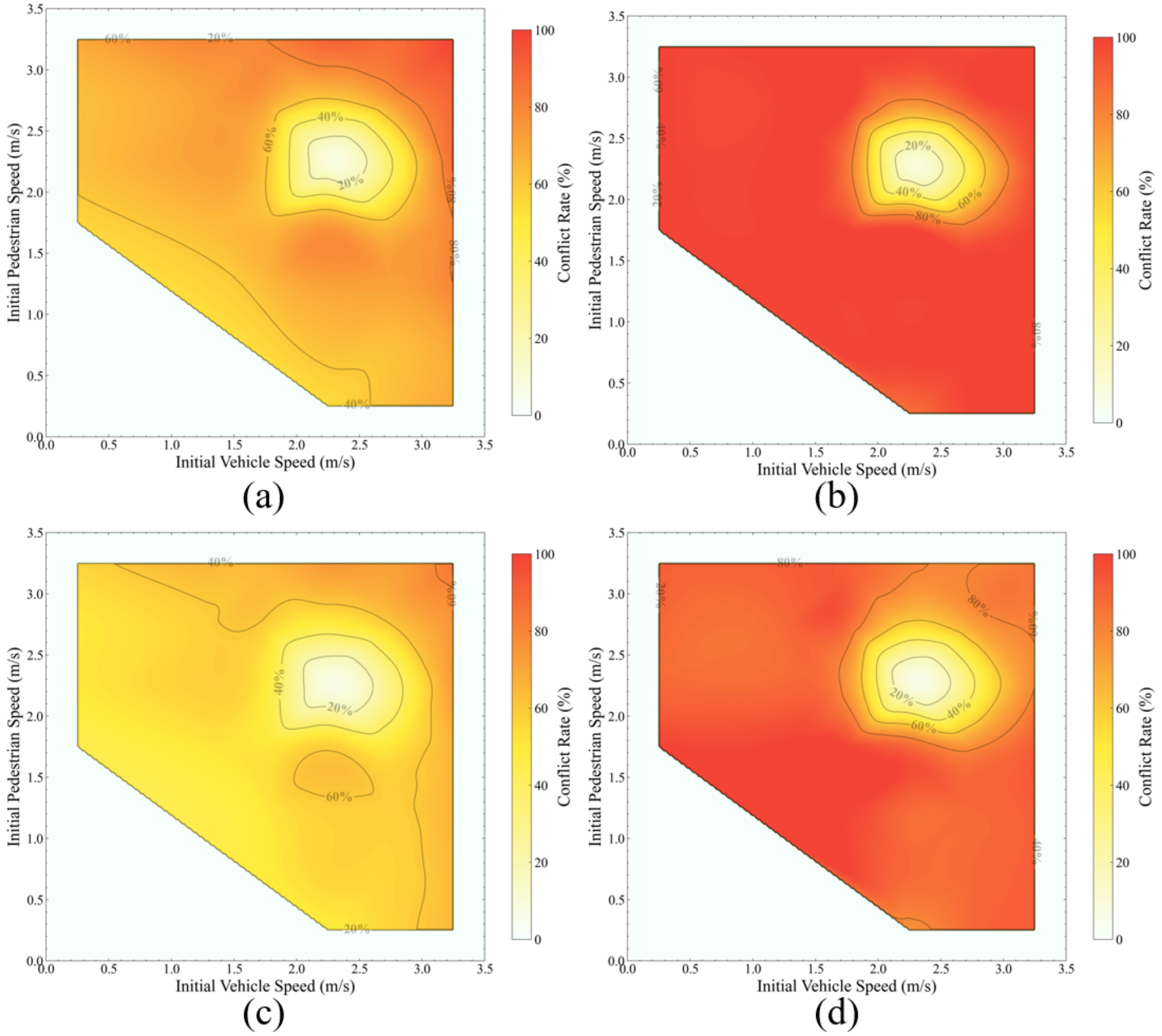

**Figure 11 : Conflict rate across different combinations of initial vehicle and pedestrian speeds: (a) three-stage framework data with CurvTTC; (b) CARLA-generated data with CurvTTC; (c) three-stage framework data with PET; (d) CARLA-generated data with PET**

As shown in Figure 11 (a), conflict rates from the three-stage framework increase with both vehicle and pedestrian speeds, falling below 40% when either agent speed remains below 1.0 m/s and exceeding 80% as both agents approach 3.0–3.5 m/s, consistent with reduced reaction time at higher speeds. In contrast, Figure 11 (b) shows that CARLA-generated data exhibits uniformly high conflict rates across most speed combinations, with an anomalous low-conflict region when both speeds fall within 1.5–2.5 m/s, contradicting real-world dynamics where higher speeds increase conflict risk. This anomaly stems from CARLA's fixed proximity-based braking thresholds, which fail to trigger timely intervention when pedestrians enter the conflict zone at high speed. The PET-based results corroborate these patterns. Figure 11 (c) reproduces the speed-dependent structure of Figure 11 (a) with overall lower conflict rates, reflecting the stricter event-based definition of PET, while Figure 11 (d) again shows uniformly elevated conflict rates for the CARLA autopilot with limited sensitivity to speed conditions. The consistency between the CurvTTC- and PET-based results indicates that the observed speed-risk relationships, and the contrast between the three-stage framework and the default autopilot, are robust to the choice of conflict indicator, with CurvTTC additionally covering the interactions resolved by evasive maneuvers before any conflict-area crossing occurred.

### 4.6. Pedestrian Yielding Behavior Analysis

This section investigates how pedestrians adapt their crossing decisions in response to approaching vehicle characteristics, examining yielding probability as a function of initial vehicle speed, acceleration, and distance at conflict onset. Results are compared against CARLA simulator outputs to assess whether the framework captures the behavioral adaptability observed in real-world pedestrian decision-making.

Similar to conflict analysis, the first frame where CurvTTC $< 5$ s was marked as the onset, when the initial vehicle speed, acceleration, and distance to the pedestrian were recorded. A pedestrian was classified as yielding if they stopped or restricted their movement in response to the approaching vehicle, rather than completing the crossing.

For comparison, vehicle–pedestrian interactions were also generated using the CARLA default controller, with initial vehicle speed, acceleration, and distance recorded at the frame where CurvTTC first dropped below 5 seconds.

As shown in Figure 12 (a), pedestrian yielding probability exceeds 60% when vehicle-pedestrian distance falls below 5 m and initial vehicle speed exceeds 2.5 m/s, confirming that pedestrians are more likely to yield when vehicles approach rapidly at close range. Figure 12 (b) further shows elevated yielding probability when the vehicle maintains positive acceleration at close range, reflecting pedestrian sensitivity to aggressive vehicle behavior. In contrast, Figure 12 (c) and (d) reveal uniformly low yielding probabilities below 20% across the entire kinematic domain, indicating that CARLA pedestrian agents do not respond to vehicle speed, acceleration, or proximity.

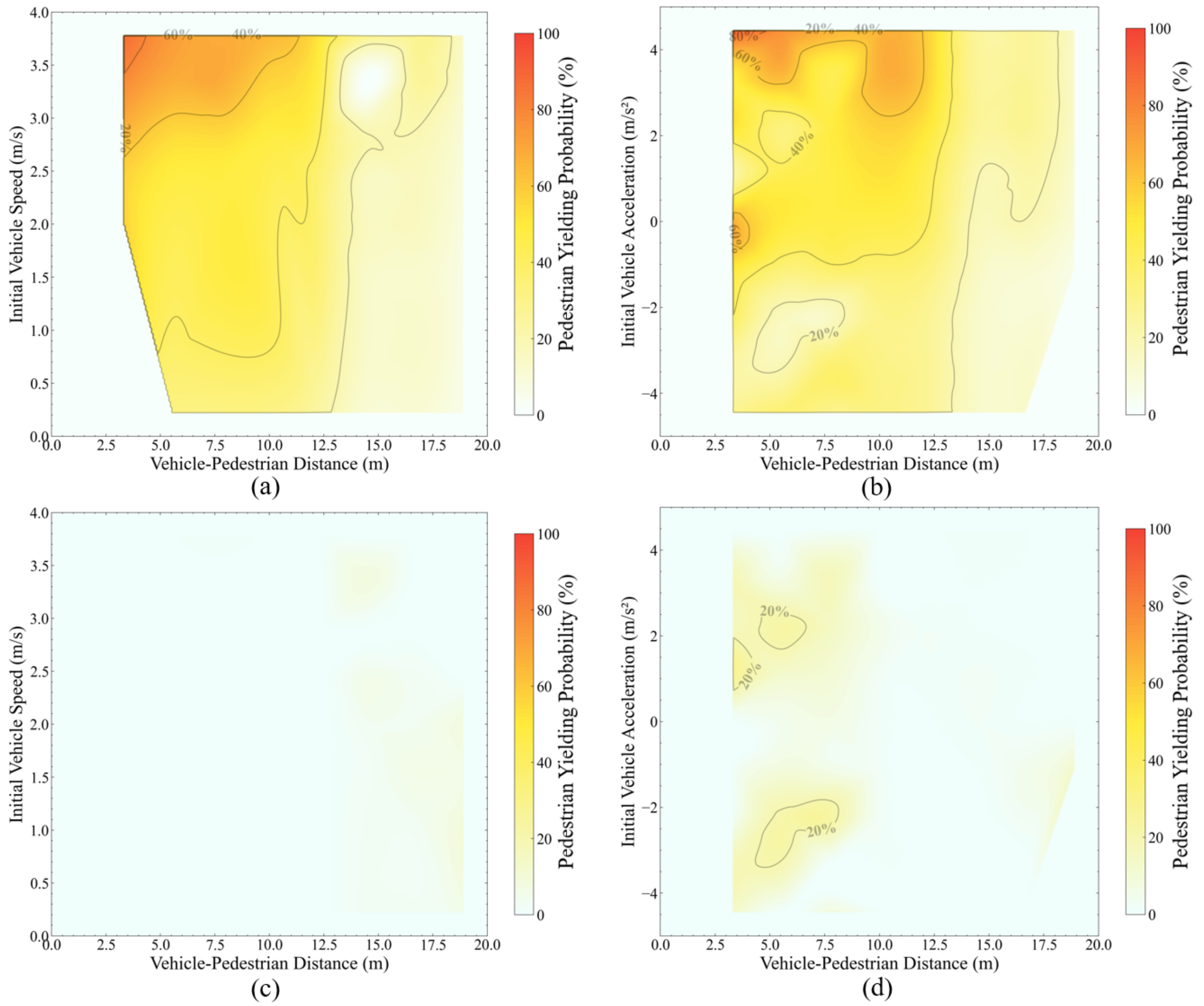

**Figure 12 : Pedestrian yielding probability contour maps as a function of initial vehicle speed/acceleration and vehicle-pedestrian distance: (a–b) three-stage framework generated data; (c–d) CARLA-generated data**

Complementing the vehicle-side view in Figure 12, a parallel analysis was conducted from the pedestrian's own kinematic perspective. As shown in Figure 13 (a) and (b), pedestrian yielding probability peaks at close vehicle–pedestrian distances when pedestrians are moving slowly or decelerating, and decays as distance increases or pedestrians accelerate forward. This is kinematically consistent with real behavior: pedestrians already slowing near an approaching vehicle are the ones who wait, while faster, accelerating pedestrians commit to crossing. In contrast, Figure 13 (c) and (d) show diffuse patterns in CARLA-generated data, with peak probabilities below 40% and no coherent dependence on pedestrian kinematics. Together with Figure 12, these results confirm that the three-stage framework reproduces context-sensitive yielding decisions responsive to both vehicle and pedestrian kinematics.

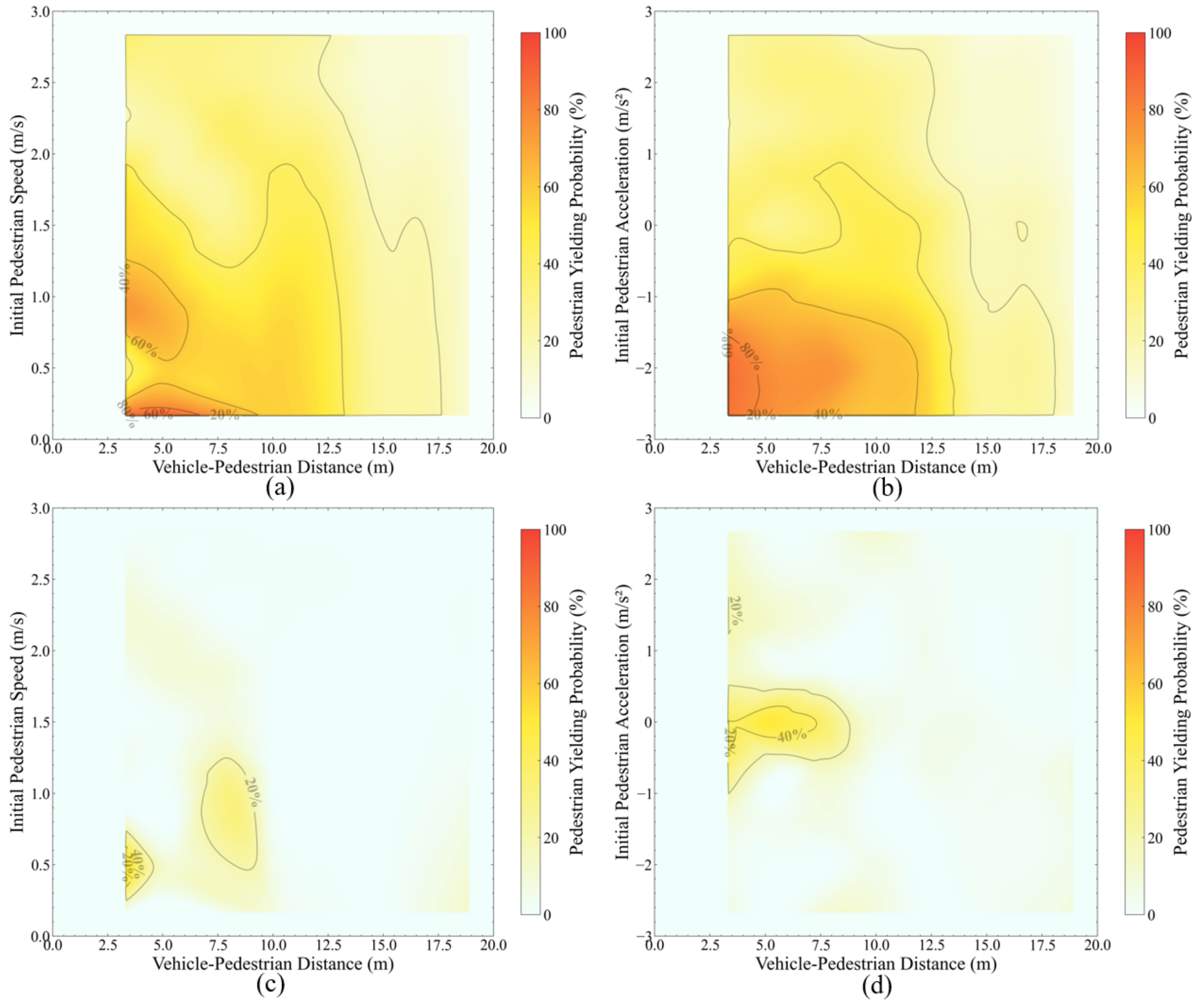

**Figure 13 : Pedestrian yielding probability contour maps as a function of initial pedestrian speed/acceleration and vehicle–pedestrian distance: (a–b) three-stage framework generated data; (c–d) CARLA-generated data**

## 4.7. Comparisons with Other Scenario Generation Frameworks

To position the proposed framework within the landscape of scenario generation research, a comparison was conducted with representative approaches. The comparison proceeds in two steps: a qualitative assessment across structural dimensions, followed by a quantitative comparison against the five frameworks for which a common evaluation protocol was methodologically feasible.

Table 12 Comparison with representative scenario generation frameworks for vehicle-pedestrian interactions

| Framework | Behavior Modeling | Real-World Data Used | Physics Engine Simulation | Large Scale Dataset Output | Quantitative Comparison |
|---|---|---|---|---|---|
| Critical pedestrian scenario library (Yang et al. 2025) | Rule-based | Yes (12 signalized intersections) | No | Partial (parameterized test-case library) | Infeasible (incompatible objective and data) |
| Human-like adversarial generation (Wang *et al.* 2026b) | Cognitive model | No | Yes (CARLA) | No | Infeasible (parameter search objective) |
| Suicidal Pedestrian (Yang et al. 2023) | Single-agent adversarial RL | No | Yes (CARLA) | No | Inappropriate (adversarial by design) |
| SceNDD++ (Ranjan et al. 2024) | Trajectory replay | Yes (instrumented vehicle, Indianapolis) | Partial (MATLAB) | Partial (MATLAB scenario files) | Infeasible (no behavior model) |
| Multi-state social force model (Yang et al. 2020) | Rule-based social force | No | No | No | ADE: 2.114; FDE: 4.298 |
| Game-theoretic predictor (Li *et al.* 2024) | Game-theoretic prediction | Yes (self-collected data) | No | No | ADE: 1.332; FDE: 2.543 |
| GT-DMIRL (Li *et al.* 2025) | Game-theoretic IRL | Yes (self-collected data) | No | No | ADE: 0.932 FDE: 2.064 |
| SF-GRUA-TF (Pang *et al.* 2025) | Social force + Transformer | Yes (self-collected data) | No | No | ADE: 0.825; FDE: 1.722 |

| Framework | Behavior Modeling | Real-World Data Used | Physics Engine Simulation | Large Scale Dataset Output | Quantitative Comparison |
|---|---|---|---|---|---|
| ResLogit pedestrian model (Al-Haideri and Farooq 2026) | Discrete choice (ResLogit) | Yes (nuScenes) | No | No | ADE: 0.821; FDE: 1.745 |
| Three-stage framework (this study) | Multi-agent RL | Yes (HDI dataset) | Yes (CARLA) | Yes (VPSCI: 198K+ episodes) | ADE: 0.072; FDE: 0.142 |

Table 12 compares frameworks along four structural dimensions. Behavior modeling ranges from trajectory replay (Ranjan et al., 2024) and fixed rules such as social force (Yang et al., 2020) and a parameterized test-case library (Yang et al., 2025), through cognitive (Wang et al., 2026), game-theoretic (Li et al., 2025; Li et al., 2024) and discrete-choice (Al-Haideri and Farooq, 2026) formulations, to learning-based models (Pang et al., 2025; Yang et al., 2023). In all of them one side of the encounter is exogenous, being replayed (Ranjan et al., 2024), scripted, or reduced to an autonomous emergency braking surrogate (Yang et al., 2025), whereas here vehicle and pedestrian co-adapt, making the conflict itself the modeled object. Real-world data underpin six of the nine frameworks, but as single-site self-collected recordings (Li et al., 2025; Li et al., 2024; Pang et al., 2025), one instrumented vehicle (Ranjan et al., 2024), twelve signalized intersections (Yang et al., 2025), or a public driving corpus (Al-Haideri and Farooq, 2026), used for offline calibration rather than closed-loop generation; the remaining three rely on assumed parameters alone (Wang et al., 2026; Yang et al., 2020; Yang et al., 2023). Physics engine simulation is present in only three frameworks, through CARLA (Wang et al., 2026; Yang et al., 2023) or partial MATLAB dynamics (Ranjan et al., 2024); without it, the trajectory models output planar point sequences lacking vehicle dynamics, occlusion, and sensor rendering, which cannot be executed as test cases. Large-scale output is rarest: only a parameterized test-case library (Yang et al., 2025) and convertible scenario files (Ranjan et al., 2024) are released, and both are configuration artifacts rather than behaviorally generated corpora. The present framework is the only entry combining all four dimensions, which is what supports scalable generation of executable safety-critical scenarios. The last column reports eligibility for the quantitative protocol, which admits a framework only if it instantiates an interactive pedestrian model aimed at naturalistic reproduction. Four fail this by design: two consume recorded trajectories as input, through resampling with a braking surrogate (Yang et al., 2025) or format conversion (Ranjan et al., 2024), while one optimizes a risk statistic (Wang et al., 2026) and one rewards collisions (Yang et al., 2023).

Five frameworks are methodologically compatible with our evaluation protocol, as each produces pedestrian trajectories that can be scored against ground truth. To ensure a fair comparison, every framework was calibrated or trained on the same HDI training set used for the proposed model rather than adopting parameters from its original publication, and all models were then evaluated on the identical 67 held-out interactions. Apart from the social force model (Yang et al. 2020), no source code was available, so each framework was re-implemented from its published description; the reported values therefore reflect these algorithms in the HDI setting, not their original results. The frameworks span rule-based, game-theoretic, inverse-reinforcement-learning, and deep-learning paradigms: the multi-state social force model (Yang et al. 2020) yielded ADE = 2.114 m and FDE = 4.298 m; the game-theoretic predictor (Li et al. 2024), 1.332 m and 2.543 m; the game-theoretic IRL model GT-DMIRL (Li et al. 2025), 0.932 m and 2.064 m; the deep social-force model SF-GRUA-TF (Pang et al. 2025), 0.825 m and 1.722 m; and the discrete-choice ResLogit model (Al-Haideri and Farooq 2026), 0.821 m and 1.745 m. The monotone decrease from the rule-based social force model to the data-driven models confirms the advantage of learning interactive behavior over fixed interaction rules. Nevertheless, the proposed Refined MA-SST-DDPG attains substantially lower error (ADE = 0.072 m, FDE = 0.142 m) than the other frameworks. Since all models were trained on the identical HDI dataset, these gains arise from the framework design rather than any data advantage, primarily owing to the multi-agent architecture and the hybrid pre-training–refinement strategy.

### 4.8. Implications for ADS Training and Testing

The VPSCI dataset addresses two distinct but complementary roles in ADS development: scenario-based testing and behavior-conditioned training.

For ADS testing, the dataset provides a structured testbed of over 198,000 safety-critical episodes across eight intersection geometries, enabling systematic evaluation of ADS collision avoidance performance under controlled but realistic conditions. Each episode records CurvTTC, agent kinematics, and conflict outcomes, allowing computation of standardized safety metrics such as conflict rate, minimum TTC, and evasion success rate. Critically, because pedestrian behaviors in VPSCI are generated by the MA-SST-DDPG policy trained on real-world HDI data rather than scripted rules, the pedestrian agent functions as a behaviorally realistic and adaptive surrogate, exposing ADS to the full range of naturalistic evasive and non-yielding responses. This enables more rigorous corner-case testing than is possible with rule-based simulators such as CARLA's default Traffic Manager.

For ADS training, the dataset supports two complementary approaches. First, the recorded vehicle and pedestrian trajectories can be used directly as imitation learning demonstrations, providing a large-scale corpus of human-like collision avoidance behaviors for behavior cloning or inverse reinforcement learning. Second, the pre-trained MA-SST-DDPG pedestrian policy can be deployed as a background agent within an ADS training environment. By fixing the pedestrian policy and treating it as part of the environment, an ADS vehicle agent can be trained against a behaviorally realistic and adaptive opponent rather than a static or scripted pedestrian. This setup directly addresses the long-standing challenge of ADS reinforcement learning, where the quality of the background agent fundamentally limits the quality of the learned driving policy. Together, these applications position the VPSCI dataset and the MA-SST-DDPG framework not merely as a data generation tool but as an integrated infrastructure for both the safety evaluation and behavioral training of ADS in safety-critical pedestrian interaction scenarios.

## 5. Conclusions

This study introduces a three-stage framework to generate the VPSCI dataset, a large-scale and realistic collection of vehicle-pedestrian safety-critical interactions. Unlike existing resources limited in scale or scenario coverage, the dataset ensures diversity across intersection types, motion states, and safety levels. Stage 1 pre-trains the MA-SST-DDPG model on real-world safety-critical data, allowing agents to learn authentic collision avoidance behaviors. Stage 2 employs online learning in CARLA, where agents adapt to diverse and safety-critical scenarios. Stage 3 uses CARLA with the Refined MA-SST-DDPG model to create a large-scale, high-quality dataset, addressing the limits of real-world data (low diversity) and simulation-only methods (low realism).

Unlike traditional methods that stop after training and directly deploy the model in simulation, the proposed three-stage framework incorporates an online learning stage and large-scale data generation. This ensures high modeling accuracy, realistic safety-critical interactions, and diversity across scenarios, behaviors, and risk levels. The Refined MA-SST-DDPG model achieves the lowest RMSE values among all compared methods, with ADE = 0.072 m and FDE = 0.142 m, outperforming the pre-trained MA-SST-DDPG model. The generated dataset comprises 198,157 episodes across eight intersection scenarios, covering 3,603 km of vehicle and 2,625 km of pedestrian trajectories. Statistical comparison between the VPSCI and HDI datasets confirmed distributional equivalence in conflict severity and close agreement in pedestrian yielding behavior, demonstrating that the synthesized data preserves both the kinematic statistics and the behavioral logic of real-world safety-critical interactions. In the Turing test with 51 participants, the realism scores of the generated interactions ranged from 3.96 to 4.08, and the paired equivalence analysis (TOST, $p < 0.001$; Cohen's $d_z = -0.016$) confirmed that they were statistically equivalent to real-world data within ±0.5 points on the 5-point scale, indicating the model's ability to replicate naturalistic evasive behaviors. Safety analysis, conducted with both CurvTTC and PET indicators, further revealed that conflict rates increase with vehicle and pedestrian speeds, and that pedestrians are more likely to cross without yielding when vehicles are decelerating or maintaining greater distance, both trends consistent with real-world behavioral logic. In contrast, CARLA-generated data failed to exhibit these patterns, highlighting the simulator's limited capacity to reproduce collision avoidance behavior. The framework was further positioned against representative scenario generation approaches through a qualitative and quantitative comparison.

The proposed framework enables the generation of large-scale safety-critical data that reflect realistic vehicle–pedestrian avoidance behaviors. It addresses two core challenges: the scarcity of real-world safety-critical data and the limited behavioral realism of simulation platforms. By learning from real interactions and deploying trained agents in simulation, the framework produces diverse and high-risk scenarios that retain human-like behavior. The VPSCI dataset serves as a benchmark resource for evaluating collision avoidance strategies, modeling pedestrian safety, and advancing behavior prediction in autonomous systems.

The dataset and framework developed in this study offer practical applications for training autonomous vehicles, improving driver-assistance systems, and enhancing the fidelity of simulation platforms. By capturing high-risk evasive interactions, the data support the development of safety-aware decision-making models. As the present evaluation focuses on safety and trajectory accuracy, future work will incorporate behavioral-diversity metrics to further quantify the coverage of rare evasive maneuvers, extend the framework to additional intersection types including T-junctions and roundabouts, incorporate multi-agent crowd scenarios and greater pedestrian behavioral diversity such as elderly or distracted pedestrians, and evaluate performance under varied environmental conditions including rain and nighttime lighting. To further close the simulation-reality gap, future research will explore the integration of world models with high-fidelity physics engines capable of reproducing vehicle dynamic responses, road surface friction, and pedestrian biomechanical kinematics. Moreover, pedestrian heading was approximated by the walking direction, and body-orientation dynamics such as yaw rate will be incorporated in future work. Additional research will investigate multi-objective reward optimization and agent-specific reward functions tailored to the distinct behavioral characteristics of vehicle and pedestrian agents, with the longer-term goal of enabling real-time deployment across broader academic and industry applications.

**Acknowledgment**

The contents of this paper present the views of the authors, who are responsible for the facts and accuracy of the data presented herein. The contents of the paper do not reflect the official views or policies of the agencies. The work was supported by NVIDIA Academic Grant “SafeSim: NVIDIA-Accelerated MARL for Generating Safety-Critical Scenarios” and by the Transportation Informatics Lab, Department of Civil & Environmental Engineering at Old Dominion University (ODU).

**References**

Akhauri, S., Zheng, L., Lin, M.C., 2020. Enhanced transfer learning for autonomous driving with systematic accident simulation. 2020 IEEE/RSJ International Conference on Intelligent Robots and Systems (IROS), 5986-5993.

Al-Haideri, R., Farooq, B.J.a.P.A., 2026. Modelling pedestrian behaviour in autonomous vehicle encounters using naturalistic dataset.

Alexiadis, V., Colyar, J., Halkias, J., Hranac, R., Mchale, G., 2004. The next generation simulation program. Institute of Transportation Engineers. ITE Journal 74 (8), 22.

Ali, Y., 2025. Autonomous vehicle lane-changing dynamics and impact on the immediate follower. Analytic Methods in Accident Research 46, 100388.

Almutairi, A., Al Asmari, A.F., Alanazi, F., Alqubaysi, T., Armghan, A., 2025. Deep learning based predictive models for real time accident prevention in autonomous vehicle networks. Scientific Reports 15 (1), 20844.

Alnowaiser, K.K., Ahmed, M.A., 2024. Steeratool: Exploiting the potential of digital twin for data generation. Internet of Things 27, 101233.

Alozi, A.R., Hussein, M., 2022. Evaluating the safety of autonomous vehicle–pedestrian interactions: An extreme value theory approach. Analytic Methods in Accident Research 35, 100230.

Alozi, A.R., Hussein, M., 2024. How do active road users act around autonomous vehicles? An inverse reinforcement learning approach. Transportation research part C: emerging technologies 161, 104572.

Anzalone, L., Barra, S., Nappi, M., 2021. Reinforced curriculum learning for autonomous driving in carla. In: Proceedings of the 2021 IEEE International Conference on Image Processing (ICIP), pp. 3318-3322.

Apostolovski, N., Trajanovski, N., Chavdar, M., Kartalov, T., Gerazov, B., Ivanovski, Z., 2022. Deep learning based multimodal information fusion for near-miss event detection in intelligent traffic monitoring systems. Complex systems: Spanning control and computational cybernetics: Applications: Dedicated to professor georgi m. Dimirovski on his anniversary. Springer, pp. 357-388.

Azfar, T., Li, J., Yu, H., Cheu, R.L., Lv, Y., Ke, R., 2024. Deep learning-based computer vision methods for complex traffic environments perception: A review. Data Science for Transportation 6 (1), 1.

Bautista-Montesano, R., Galluzzi, R., Ruan, K., Fu, Y., Di, X., 2022. Autonomous navigation at unsignalized intersections: A coupled reinforcement learning and model predictive control approach. Transportation Research Part C: Emerging Technologies 139, 103662.

Berger, V.W., Zhou, Y., 2014. Kolmogorov–smirnov test: Overview. Wiley statsref: Statistics reference online.

Bock, J., Krajewski, R., Moers, T., Runde, S., Vater, L., Eckstein, L., 2020. The ind dataset: A drone dataset of naturalistic road user trajectories at german intersections. In: Proceedings of the 2020 IEEE Intelligent Vehicles Symposium (IV), pp. 1929-1934.

Caesar, H., Bankiti, V., Lang, A.H., Vora, S., Liong, V.E., Xu, Q., Krishnan, A., Pan, Y., Baldan, G., Beijbom, O., 2020. Nuscenes: A multimodal dataset for autonomous driving. In: Proceedings of the Proceedings of the IEEE/CVF conference on computer vision and pattern recognition, pp. 11621-11631.

Chang, W.-J., Pittaluga, F., Tomizuka, M., Zhan, W., Chandraker, M., 2023. Safe-sim: Safety-critical closed-loop traffic simulation with diffusion-controllable adversaries. In: Proceedings of the European conference on computer vision, pp. 242-258.

Dave, K.M., 2024. An iot-based approach of synthetic data generation with reduced reality gap. University of Windsor (Canada).

Dieter, T.R., Weinmann, A., Jäger, S., Brucherseifer, E., 2023. Quantifying the simulation–reality gap for deep learning-based drone detection. Electronics 12 (10), 2197.

Ding, W., Xu, C., Arief, M., Lin, H., Li, B., Zhao, D., 2023. A survey on safety-critical driving scenario generation—a methodological perspective. IEEE Transactions on Intelligent Transportation Systems 24 (7), 6971-6988.

Dosovitskiy, A., Ros, G., Codevilla, F., Lopez, A., Koltun, V., 2017. Carla: An open urban driving simulator. In: Proceedings of the Conference on robot learning, pp. 1-16.

Dulac-Arnold, G., Mankowitz, D., Hester, T., 2019. Challenges of real-world reinforcement learning. arXiv preprint arXiv:1904.12901.

Feng, M., Zhao, J., Hou, C., Nie, C., Hou, J., 2025. Investigating the safety influence path of right-turn configurations on vehicle–pedestrian conflict risk at signalized intersections. Accident Analysis & Prevention 211, 107910.

Fu, T., Miranda-Moreno, L., Saunier, N.J.T.R.R., 2016. Pedestrian crosswalk safety at nonsignalized crossings during nighttime: Use of thermal video data and surrogate safety measures. 2586 (1), 90-99.

Gaidon, A., Wang, Q., Cabon, Y., Vig, E., 2016. Virtual worlds as proxy for multi-object tracking analysis. In: Proceedings of the Proceedings of the IEEE conference on computer vision and pattern recognition, pp. 4340-4349.

Goulet-Pelletier, J.-C., Cousineau, D., 2018. A review of effect sizes and their confidence intervals, part i: The cohen'sd family. The Quantitative Methods for Psychology 14 (4), 242-265.

Grislain, C., Vuorio, R., Lu, C., Whiteson, S., 2024. Igdrivsim: A benchmark for the imitation gap in autonomous driving. arXiv preprint arXiv:2411.04653.

Gu, A., Dao, T., 2023. Mamba: Linear-time sequence modeling with selective state spaces. arXiv preprint arXiv:2312.00752.

Guo, H., Xie, K., Keyvan-Ekbatani, M., 2023. Modeling driver's evasive behavior during safety–critical lane changes: Two-dimensional time-to-collision and deep reinforcement learning. Accident Analysis & Prevention 186, 107063.

Gupta, S., Zaki, M.H., Vela, A., 2022. Generative modeling of pedestrian behavior: A receding horizon optimization-based trajectory planning approach. IEEE Access 10, 81624-81641.

Haarnoja, T., Zhou, A., Abbeel, P., Levine, S., 2018. Soft actor-critic: Off-policy maximum entropy deep reinforcement learning with a stochastic actor. In: Proceedings of the International conference on machine learning, pp. 1861-1870.

Han, K., Wang, Y., Chen, H., Chen, X., Guo, J., Liu, Z., Tang, Y., Xiao, A., Xu, C., Xu, Y., 2022. A survey on vision transformer. IEEE transactions on pattern analysis and machine intelligence 45 (1), 87-110.

Harkin, K.A., Harkin, A.M., Gögel, C., Schade, J., Petzoldt, T., 2024. How do vulnerable road users evaluate automated vehicles in urban traffic? A focus group study with pedestrians, cyclists, e-scooter riders, older adults, and people with walking disabilities. Transportation Research Part F: Traffic Psychology and Behaviour 104, 59-71.

He, Z., 2017. Research based on high-fidelity ngsim vehicle trajectory datasets: A review. Research Gate, 1-33.

Houston, J., Zuidhof, G., Bergamini, L., Ye, Y., Chen, L., Jain, A., Omari, S., Iglovikov, V., Ondruska, P., 2020. One thousand and one hours: Self-driving motion prediction dataset. In: Proceedings of the Conference on Robot Learning, pp. 409-418.

Hu, J., Wang, Y., Cheng, S., Xu, J., Wang, N., Fu, B., Ning, Z., Li, J., Chen, H., Feng, C., 2025. A survey of decision-making and planning methods for self-driving vehicles. Frontiers in Neurorobotics 19, 1451923.

Hu, X., Li, S., Huang, T., Tang, B., Huai, R., Chen, L., 2023. How simulation helps autonomous driving: A survey of sim2real, digital twins, and parallel intelligence. IEEE Transactions on Intelligent Vehicles 9, 593-612.

Huang, X., Cheng, X., Geng, Q., Cao, B., Zhou, D., Wang, P., Lin, Y., Yang, R., 2018. The apolloscape dataset for autonomous driving. In: Proceedings of the Proceedings of the IEEE conference on computer vision and pattern recognition workshops, pp. 954-960.

Imaseki, T., Sugasawa, F., Kawakami, E., Mouri, H., 2023. Criticality metrics study for safety evaluation of merge driving scenarios, using near-miss video data. SAE International Journal of Transportation Safety 12 (09-12-01-0002), 25-42.

Islam, Z., Abdel-Aty, M., Goswamy, A., Abdelraouf, A., Zheng, O.J.S.R., 2023. Effect of signal timing on vehicles' near misses at intersections. 13 (1), 9065.

Ji, D., Zhang, Z., Zhao, M., He, Z., Guo, Z., 2025. A multi-objective adaptive memetic algorithm and engineering application for a double-floor layout problem with separate human and vehicle transport elevators. Engineering Applications of Artificial Intelligence 155, 111010.

Jian, Z., Qingxia, L., Shengde, D., Ronggui, Z., 2018. Risk assessment method of weaving area based on traffic conflict of ttc. In: Proceedings of the Proceedings of the Asia-Pacific Conference on Intelligent Medical 2018 & International Conference on Transportation and Traffic Engineering 2018, pp. 95-98.

Kamal, H., Yánez, W., Hassan, S., Sobhy, D., 2024. Digital-twin-based deep reinforcement learning approach for adaptive traffic signal control. IEEE Internet of Things Journal 11 (12), 21946-21953.

Khan, M.A., Karim, M.R., Kim, Y., 2018. A two-stage big data analytics framework with real world applications using spark machine learning and long short-term memory network. Symmetry 10 (10), 485.

Khuzam, E.A., Lanzaro, G., Sayed, T., 2025. Impact of jaywalking on pedestrian interaction behavior: A multiagent markov game-based analysis. Accident Analysis & Prevention 220, 108141.

Koo, T.K., Li, M.Y., 2016. A guideline of selecting and reporting intraclass correlation coefficients for reliability research. Journal of chiropractic medicine 15 (2), 155-163.
Krajewski, R., Bock, J., Kloeker, L., Eckstein, L., 2018. The highd dataset: A drone dataset of naturalistic vehicle trajectories on german highways for validation of highly automated driving systems. In: Proceedings of the 2018 21st international conference on intelligent transportation systems (ITSC), pp. 2118-2125.
Krajewski, R., Moers, T., Bock, J., Vater, L., Eckstein, L., 2020. The round dataset: A drone dataset of road user trajectories at roundabouts in germany. In: Proceedings of the 2020 IEEE 23rd International Conference on Intelligent Transportation Systems (ITSC), pp. 1-6.
Krajzewicz, D., 2010. Traffic simulation with sumo–simulation of urban mobility. Fundamentals of traffic simulation, 269-293.
Lanzaro, G., Sayed, T., 2025. Evaluating driver-pedestrian interaction behavior in different environments via markov-game-based inverse reinforcement learning. Expert Systems with Applications 260, 125405.
Lauzon, C., Caffo, B., 2009. Easy multiplicity control in equivalence testing using two one-sided tests. The American Statistician 63 (2), 147-154.
Li, K., Zhang, R., Wang, H., Yu, F., 2021. Multi-intelligent connected vehicle longitudinal collision avoidance control and exhaust emission evaluation based on parallel theory. Process Safety and Environmental Protection 150, 259-268.
Li, W., Li, X., Li, L., Tang, Y., Hu, Y.J.a.A., Prevention, 2025. Simulation of human–vehicle interaction at right-turn unsignalized intersections: A game-theoretic deep maximum entropy inverse reinforcement learning method. 214, 107960.
Li, W., Pan, C.W., Zhang, R., Ren, J.P., Ma, Y.X., Fang, J., Yan, F.L., Geng, Q.C., Huang, X.Y., Gong, H.J., Xu, W.W., Wang, G.P., Manocha, D., Yang, R.G., 2019. Aads: Augmented autonomous driving simulation using data-driven algorithms. Science Robotics 4 (28), eaaw0863.
Li, W., Zhang, Y., Li, L., Lv, Y., Wang, M.J.I.T.O.I.T.S., 2024. A pedestrian trajectory prediction model for right-turn unsignalized intersections based on game theory. 25 (8), 9643-9658.
Li, Y., Liu, F., Xing, L., He, Y., Dong, C., Yuan, C., Chen, J., Tong, L., 2023. Data generation for connected and automated vehicle tests using deep learning models. Accident Analysis & Prevention 190, 107192.
Lillicrap, T., 2015. Continuous control with deep reinforcement learning. arXiv preprint arXiv:1509.02971.
Lin, H., Huang, X., Phan, T., Hayden, D., Zhang, H., Zhao, D., Srinivasa, S., Wolff, E., Chen, H., 2024. Causal composition diffusion model for closed-loop traffic generation. In: Proceedings of the Proceedings of the IEEE/CVF Conference on Computer Vision and Pattern Recognition, pp. 27542-27552.
Lowe, R., Wu, Y.I., Tamar, A., Harb, J., Pieter Abbeel, O., Mordatch, I., 2017. Multi-agent actor-critic for mixed cooperative-competitive environments. Advances in neural information processing systems 30.
Luo, M., Li, H., Luo, W., Li, H., Li, J., 2025. Goal-oriented autonomous decision-making for social robots via collaborative interactive inverse reinforcement learning approach. Scientific Reports 15 (1), 27724.
Minderhoud, M.M., Bovy, P.H., 2001. Extended time-to-collision measures for road traffic safety assessment. Accident Analysis & Prevention 33 (1), 89-97.
Nasernejad, P., Sayed, T., Alsaleh, R., 2023. Multiagent modeling of pedestrian-vehicle conflicts using adversarial inverse reinforcement learning. Transportmetrica A: Transport Science 19 (3), 2061081.
Ni, Y., Wang, M., Sun, J., Li, K., 2016. Evaluation of pedestrian safety at intersections: A theoretical framework based on pedestrian-vehicle interaction patterns. Accident Analysis & Prevention 96, 118-129.

Pang, T., Chen, H., Wang, Z., Zhang, X.J.D.S.P., 2025. A data-driven social force-gru attention-transformer for pedestrian trajectory prediction. 105670.

Pérez-Gil, Ó., Barea, R., López-Guillén, E., Bergasa, L.M., Gomez-Huelamo, C., Gutiérrez, R., Diaz-Diaz, A., 2022. Deep reinforcement learning based control for autonomous vehicles in carla. Multimedia Tools and Applications 81 (3), 3553-3576.

Pradana, H., Dao, M.-S., Zettsu, K., 2022. Augmenting ego-vehicle for traffic near-miss and accident classification dataset using manipulating conditional style translation. In: Proceedings of the 2022 International Conference on Digital Image Computing: Techniques and Applications (DICTA), pp. 1-8.

Pu, Q., Xie, K., Guo, H., 2026a. Modeling interactive car-following behaviors of automated and human-driven vehicles in safety-critical events: A multi-agent state-space attention-enhanced framework. Accident Analysis & Prevention 229, 108447.

Pu, Q., Xie, K., Guo, H., Zhu, Y., 2025a. Modeling crash avoidance behaviors in vehicle-pedestrian near-miss scenarios: Curvilinear time-to-collision and mamba-driven deep reinforcement learning. Accident Analysis & Prevention 214, 107984.

Pu, Q., Xie, K., Guo, H., Zhu, Y., 2026b. Modeling interactive crash avoidance behaviors: A multi-agent state-space transformer-enhanced reinforcement learning framework. Accident Analysis & Prevention 226, 108334.

Pu, Q., Xie, K., Liu, Y., 2026c. Vlm-vpi: A vision-language reasoning framework for improving automated vehicle-pedestrian interactions. arXiv preprint arXiv:2604.23934.

Pu, Q., Xie, K., Yang, H., Zhai, G., 2026d. A vision-and-knowledge enhanced large language model for generalizable pedestrian crossing behavior inference. arXiv preprint arXiv:2601.00694.

Pu, Q., Zhu, Y., Wang, J., Yang, H., Xie, K., Cui, S., 2025b. Drone data analytics for measuring traffic metrics at intersections in high-density areas. Transportation Research Record, 03611981241311566.

Quispe, H., Sumire, J., Condori, P., Alvarez, E., Vera, H., 2022. I see you: A vehicle-pedestrian interaction dataset from traffic surveillance cameras. arXiv preprint arXiv:2211.09342.

Ranjan, N., Prabu, A., He, Z., Li, L., 2024. Scendd++: An enhanced scenario-based naturalistic driving dataset. In: Proceedings of the 2024 International Conference on Cyber-Physical Social Intelligence (ICCSI), pp. 1-6.

Rempe, D., Philion, J., Guibas, L.J., Fidler, S., Litany, O., 2021. Generating useful accident-prone driving scenarios via a learned traffic prior. In: Proceedings of the Proceedings of the IEEE/CVF Conference on Computer Vision and Pattern Recognition, pp. 17305-17315.

Richter, S.R., Vineet, V., Roth, S., Koltun, V., 2016. Playing for data: Ground truth from computer games. In: Proceedings of the Computer Vision–ECCV 2016: 14th European Conference, Amsterdam, The Netherlands, October 11-14, 2016, Proceedings, Part II 14, pp. 102-118.

Ros, G., Sellart, L., Materzynska, J., Vazquez, D., Lopez, A.M., 2016. The synthia dataset: A large collection of synthetic images for semantic segmentation of urban scenes. In: Proceedings of the Proceedings of the IEEE conference on computer vision and pattern recognition, pp. 3234-3243.

Schulman, J., Wolski, F., Dhariwal, P., Radford, A., Klimov, O., 2017. Proximal policy optimization algorithms. arXiv preprint arXiv:1707.06347.

Shi, J., Zhang, T., Zhan, J., Chen, S., Xin, J., Zheng, N., 2023. Efficient lane-changing behavior planning via reinforcement learning with imitation learning initialization. In: Proceedings of the 2023 IEEE Intelligent Vehicles Symposium (IV), pp. 1-8.

Song, Q., Bensoussan, A., Mousavi, M.R., 2025. Synthetic versus real: An analysis of critical scenarios for autonomous vehicle testing. Automated Software Engineering 32 (2), 37.

Sun, P., Kretzschmar, H., Dotiwalla, X., Chouard, A., Patnaik, V., Tsui, P., Guo, J., Zhou, Y., Chai, Y., Caine, B., 2020. Scalability in perception for autonomous driving: Waymo open dataset. In:

Proceedings of the Proceedings of the IEEE/CVF conference on computer vision and pattern recognition, pp. 2446-2454.
Suo, S., Regalado, S., Casas, S., Urtasun, R., 2021. Trafficsim: Learning to simulate realistic multi-agent behaviors. In: Proceedings of the Proceedings of the IEEE/CVF Conference on Computer Vision and Pattern Recognition, pp. 10400-10409.
Suzuki, T., Aoki, Y., Kataoka, H., 2017. Pedestrian near-miss analysis on vehicle-mounted driving recorders. In: Proceedings of the 2017 Fifteenth IAPR International Conference on machine vision applications (MVA), pp. 416-419.
Talebpour, A., Mahmassani, H.S., Hamdar, S.H., 2024. Third generation simulation data (tgsim): A closer look at the impacts of automated driving systems on human behavior. United States. Department of Transportation. Intelligent Transportation ….
Tesla, I., 2024. Tesla model 3 owner's manual.
Tian, D., Zhou, J., Han, X., Lang, P., 2024. Robust platoon control of mixed autonomous and human-driven vehicles for obstacle collision avoidance: A cooperative sensing-based adaptive model predictive control approach. Engineering 42, 244-266.
Tselentis, D.I., Papadimitriou, E., 2023. Time-series clustering for pattern recognition of speed and heart rate while driving: A magnifying lens on the seconds around harsh events. Transportation research part F: traffic psychology and behaviour 98, 254-268.
Wang, C., Guo, F., Zhao, S., Zhu, Z., Zhang, Y., 2024. Safety assessment for autonomous vehicles: A reference driver model for highway merging scenarios. Accident Analysis & Prevention 206, 107710.
Wang, X., Wang, S., Liang, X., Zhao, D., Huang, J., Xu, X., Dai, B., Miao, Q., 2022. Deep reinforcement learning: A survey. IEEE Transactions on Neural Networks and Learning Systems 35 (4), 5064-5078.
Wang, Y., Dogar, M., Darling, R., Markkula, G., 2026a. Realistic adversarial scenario generation via human-like pedestrian model for autonomous vehicle control parameter optimisation. arXiv preprint arXiv:2601.02082.
Wang, Y., Dogar, M., Darling, R., Markkula, G.J.a.P.A., 2026b. Realistic adversarial scenario generation via human-like pedestrian model for autonomous vehicle control parameter optimisation.
Wilson, B., Qi, W., Agarwal, T., Lambert, J., Singh, J., Khandelwal, S., Pan, B., Kumar, R., Hartnett, A., Pontes, J.K., 2023. Argoverse 2: Next generation datasets for self-driving perception and forecasting. arXiv preprint arXiv:2301.00493.
Wrenninge, M., Unger, J., 2018. Synscapes: A photorealistic synthetic dataset for street scene parsing. arXiv preprint arXiv:1810.08705.
Xiao, B., Feng, C., Huang, Z., Yan, F., Zhong, Y., Ma, L., 2025. Robotron-sim: Improving real-world driving via simulated hard-case.
Xu, C., Gao, J., Zuo, F., Ozbay, K., 2024. Estimating urban traffic safety and analyzing spatial patterns through the integration of city-wide near-miss data: A new york city case study. Applied Sciences (2076-3417) 14 (14).
Xu, C., Petiushko, A., Zhao, D., Li, B., 2017. Diffscene: Diffusion-based safety-critical scenario generation for autonomous vehicles. In: Proceedings of the Proceedings of the AAAI conference on artificial intelligence, pp. 8797-8805.
Xu, D., Chen, Y., Ivanovic, B., Pavone, M., 2023. Bits: Bi-level imitation for traffic simulation. In: Proceedings of the 2023 IEEE International Conference on Robotics and Automation (ICRA), pp. 2929-2936.
Yan, M., Xiong, R., Wang, Y., Li, C., 2023. Edge computing task offloading optimization for a uav-assisted internet of vehicles via deep reinforcement learning. IEEE Transactions on Vehicular Technology 73 (4), 5647-5658.

Yang, D., Redmill, K., Özgüner, Ü., 2020. A multi-state social force based framework for vehicle-pedestrian interaction in uncontrolled pedestrian crossing scenarios. In: Proceedings of the 2020 IEEE Intelligent Vehicles Symposium (IV), pp. 1807-1812.

Yang, L., Liu, S., Feng, S., Wang, H., Zhao, X., Qu, G., Fang, S.J.a.A., Prevention, 2025. Generation of critical pedestrian scenarios for autonomous vehicle testing. 214, 107962.

Yang, Y., Kujanpää, K., Babadi, I.A., Pajarinen, J., Ilin, A., 2023. Suicidal pedestrian: Generation of safety-critical scenarios for autonomous vehicles. In: Proceedings of the 2023 IEEE 26th International Conference on Intelligent Transportation Systems (ITSC), pp. 1983-1988.

Yang, Z., Li, Z., Hu, J., Zhang, Y., 2024. Dynamically expanding capacity of autonomous driving with near-miss focused training framework. In: Proceedings of the International Conference on Transportation and Development 2024, pp. 616-626.

Yao, R., Sun, X., 2025. Hierarchical prediction uncertainty-aware motion planning for autonomous driving in lane-changing scenarios. Transportation Research Part C: Emerging Technologies 171, 104962.

Yin, H., Chen, J., Pan, S.J., Tschiatschek, S., 2021. Sequential generative exploration model for partially observable reinforcement learning. In: Proceedings of the Proceedings of the aaai conference on artificial intelligence, pp. 10700-10708.

Zhan, W., Sun, L., Wang, D., Shi, H., Clausse, A., Naumann, M., Kummerle, J., Konigshof, H., Stiller, C., De La Fortelle, A., 2019. Interaction dataset: An international, adversarial and cooperative motion dataset in interactive driving scenarios with semantic maps. arXiv preprint arXiv:1910.03088.

Zhang, J., Xu, C., Li, B., 2024. Chatscene: Knowledge-enabled safety-critical scenario generation for autonomous vehicles. In: Proceedings of the Proceedings of the IEEE/CVF Conference on Computer Vision and Pattern Recognition, pp. 15459-15469.

Zhang, X., Zheng, Y., Wang, L., Abdulali, A., Iida, F., 2023. Multi-agent collaborative target search based on the multi-agent deep deterministic policy gradient with emotional intrinsic motivation. Applied Sciences 13 (21), 11951.

Zhang, Z., Trivedi, C., Liu, X., 2018. Automated detection of grade-crossing-trespassing near misses based on computer vision analysis of surveillance video data. Safety science 110, 276-285.

Zhao, J., Lv, Y., Zeng, Q., Wan, L., 2022. Online policy learning-based output-feedback optimal control of continuous-time systems. IEEE Transactions on Circuits and Systems II: Express Briefs 71 (2), 652-656.

Zheng, S., Liu, H., 2019. Improved multi-agent deep deterministic policy gradient for path planning-based crowd simulation. Ieee Access 7, 147755-147770.

Zhou, R., Huang, H., Zhang, G., Zhou, H., Bian, J., 2025. Crash-based safety testing of autonomous vehicles: Insights from generating safety-critical scenarios based on in-depth crash data. IEEE Transactions on Intelligent Transportation Systems.

Zuo, J., Hu, H., Zhou, Z., Cui, Y., Liu, Z., Wang, J., Guan, N., Wang, J., Xue, C.J., 2025. Ralad: Bridging the real-to-sim domain gap in autonomous driving with retrieval-augmented learning. ArXiv abs/2501.12296.